%% file: main.tex
\pdfoutput=1
\documentclass[10pt,twocolumn,letterpaper]{article}
\usepackage{paperstyle}

\input{preamble}
\definecolor{linkblue}{rgb}{0.21,0.49,0.74}
\usepackage[pagebackref,breaklinks,colorlinks,allcolors=linkblue]{hyperref}
\title{Events as Roots: Error-Controlled Continuous-Time Event Simulation\\in Gaussian Driving Scenes}
\author{Siying Li \quad Yuhang Chen \quad An Xie \quad Yujia Zhao \quad Yuanbiao Cao\\
Ying Ni$^{*}$ \quad Jie Sun \quad Jian Sun \quad Haotian Shi$^{*}$\\[0.4em]
College of Transportation, Tongji University}
\begin{document}
\twocolumn[{%
\renewcommand\twocolumn[1][]{#1}%
\maketitle
\begin{center}
    \centering
    \includegraphics[width=\linewidth]{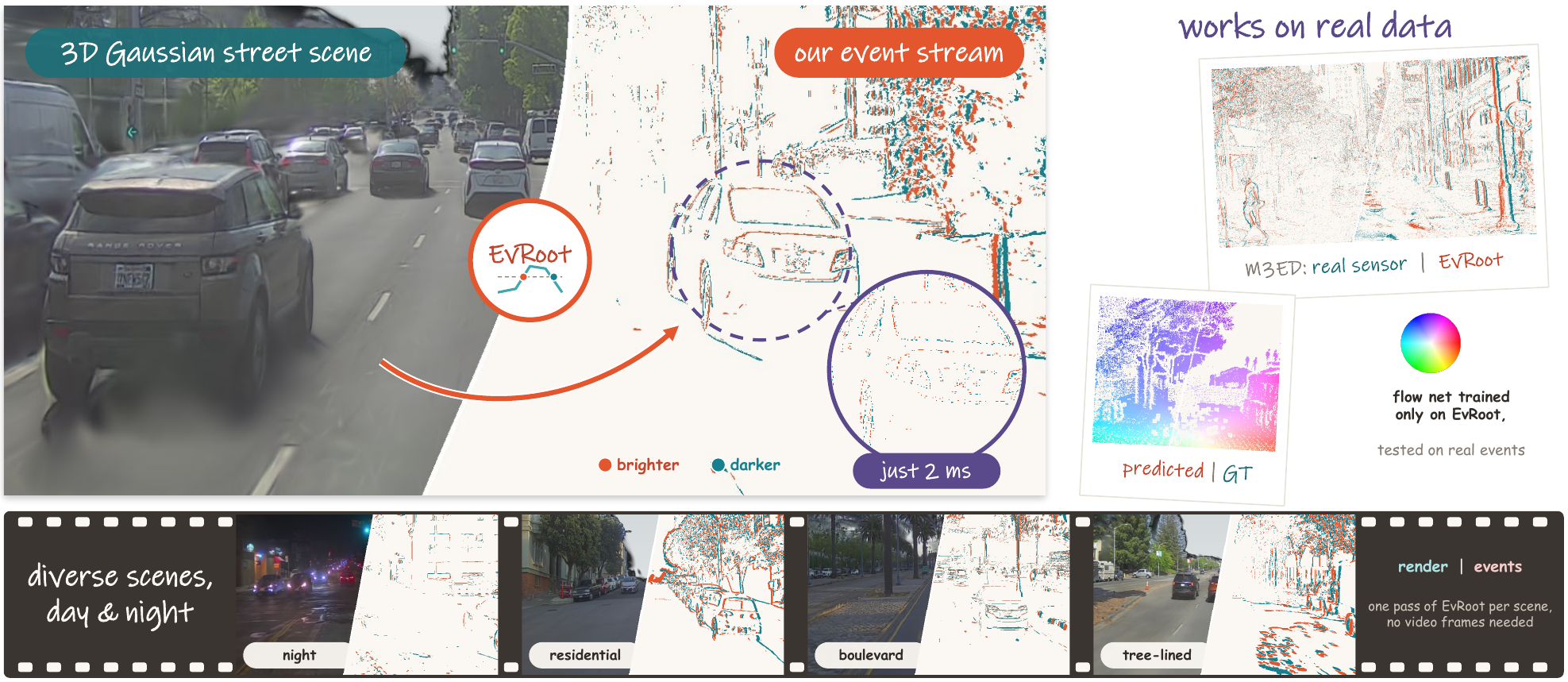}
    \vspace{-1.5em}
    \captionof{figure}{\textbf{EvRoot solves event times on Gaussian street scenes instead of sampling them from frames.}
    Left: a rendering of a reconstructed street scene (left half) and the asynchronous event stream that EvRoot computes from it (right half); the inset shows that only 2\,ms of events already outline a moving vehicle.
    Right: on real recordings, EvRoot events resemble those of a physical sensor, and an optical-flow network trained only on EvRoot events transfers to real events.
    Bottom: EvRoot applies to diverse driving scenes.}
    \label{fig:teaser}
\end{center}
\vspace{1.0em}
}]
{\renewcommand{\thefootnote}{*}\footnotetext{Corresponding authors.}}

\input{sec/0_abstract}    
\input{sec/1_intro}
\input{sec/2_related}
\input{sec/3_method}
\input{sec/4_experiments}
\input{sec/5_conclusion}
\input{sec/6_llm}

{
    \small
    \bibliographystyle{refstyle}
    \bibliography{main}
}

\input{sec/X_suppl}

\end{document}

%% file: preamble.tex
\usepackage{xspace}
\newcommand{\method}{EvRoot\xspace}

\usepackage{colortbl}
\usepackage{xcolor}
\definecolor{rkA}{HTML}{F4C2C2}   %
\definecolor{rkB}{HTML}{FBE0CC}   %
\definecolor{rkC}{HTML}{FEF4D8}   %
\definecolor{abl}{HTML}{B03A48}   %
\newcommand{\fst}[1]{\cellcolor{rkA}\textbf{#1}}
\newcommand{\snd}[1]{\cellcolor{rkB}#1}
\newcommand{\trd}[1]{\cellcolor{rkC}#1}
\usepackage{multirow}
\newcommand{\dg}[1]{{\scriptsize\textcolor{abl}{$\times$#1}}}
\usepackage{algorithm}
\usepackage{algpseudocode}

%% file: sec/0_abstract.tex
\begin{abstract}
Compared with frame cameras, event cameras offer higher temporal resolution and a wider dynamic range for autonomous driving, yet acquiring and annotating real event data remains costly. Existing simulators typically estimate event times from discrete brightness samples, tying timing accuracy to sampling density while incurring substantial rendering costs and errors at occlusion boundaries.
We present \method{}, a continuous-time event simulator for driving scenes represented by 3D Gaussians that formulates event generation as root finding for the brightness-threshold equation under explicit error control.
An analytic tangent renderer computes log brightness and its temporal derivative in a single pass.
An embedded Hermite controller adaptively refines time intervals until coarse and refined interpolants agree on the number, polarity, and timing of threshold crossings within a prescribed time tolerance.
Visibility-aware bracketing localizes events at occlusion boundaries, while sparse tile scheduling restricts rendering to regions requiring refinement.
On synthetic driving benchmarks, \method{} achieves timing accuracy comparable to 64\,kHz frame-driven simulation using approximately $1/100$ of the rendering cost and running $5\times$ faster.
Evaluations on real recordings demonstrate improved event fidelity, while optical-flow networks trained on our simulated events perform comparably to those trained on real events.
Code is available at \url{https://github.com/siyingli99/EvRoot}.
\end{abstract}

%% file: sec/1_intro.tex
\section{Introduction}
\label{sec:intro}

Autonomous driving demands reliable perception under fast motion, abrupt illumination changes, and sudden occlusions, conditions in which conventional frame cameras suffer from motion blur, limited dynamic range, and latency.
Event cameras address these limitations by asynchronously reporting per-pixel log-brightness changes with microsecond resolution and high dynamic range~\cite{lichtsteiner2008dvs,gallego2022event}.
Exploiting these advantages, however, requires large amounts of training data, whereas real event recordings~\cite{zhu2018mvsec,gehrig2021dsec,chaney2023m3ed} demand dedicated sensors, careful calibration, and costly annotation.
Event simulation has therefore become an important means of producing data for event-based perception.
A practical event simulator should reproduce precise event timing, which is the defining property of event data, remain faithful in the presence of moving objects and occlusions, and scale efficiently to large driving scenes.

Existing simulators can be categorized by how they determine event times.
\emph{Frame-driven} simulators~\cite{rebecq2018esim,gehrig2020vid2e,hu2021v2e,joubert2021icns,lin2022dvsvoltmeter} interpolate log brightness between consecutive video frames to locate threshold crossings.
\emph{Learning-based} simulators~\cite{zhu2021eventgan,zhang2024v2ce} predict event representations from frames with neural networks.
\emph{Reconstruction-based} simulators exploit driving scenes reconstructed with 3D Gaussian Splatting (3DGS)~\cite{kerbl2023gaussian,yan2024streetgaussians,zhou2024drivinggaussian,chen2025omnire}, in which brightness can be rendered at arbitrary times.
GS2E~\cite{li2025gs2e} renders dense frames for a frame-driven event model, whereas TIDES~\cite{thirgood2026tides} predicts crossings from analytic temporal derivatives and shortens its steps where a visibility risk score is high.
Across these paradigms, event times are obtained as a by-product of sampling or prediction, and none of these simulators estimate the timing error of the events they emit.

This sampling-based view leaves three challenges unresolved.
First, timing accuracy cannot be controlled directly.
The error of each event is dictated by the frame rate or step size, and even adaptive steps are chosen from heuristic risk scores, making accuracy tightly coupled with cost, with its actual level being unknown.
Second, occlusions violate the smoothness that interpolation relies on.
When an object boundary sweeps across a pixel, its brightness bends or jumps precisely where many events are triggered; interpolation misplaces these events, while suppressing them discards genuine boundary events.
Third, computation is misaligned with the distribution of events.
Events are sparse in space and time, yet existing simulators typically render the full image at every frame or step, which scales poorly to complete driving scenes.

The key insight of this work is that event times should be \emph{solved rather than sampled}.
For an ideal sensor, each event time is a root of $L(\mathbf{x},t)-L_{\mathrm{ref}}(\mathbf{x})=\pm C$, where $L$ denotes the log brightness of pixel $\mathbf{x}$, $L_{\mathrm{ref}}$ its reference level, and $C$ the contrast threshold.
Since a Gaussian scene provides both $L$ and its temporal derivative in closed form, event simulation can be cast as root finding with explicit error control, in the spirit of event location in numerical ODE solvers~\cite{hairer1993solving,shampine2000event}.
This formulation addresses the three challenges in turn: an embedded error estimate turns the time tolerance into an explicit parameter; occlusions are observable inside the renderer and can be localized in time; and brightness bounds exclude crossings for most pixels, confining refinement to a small set of tiles and intervals.

Building on this insight, we propose \method, a continuous-time event simulator for Gaussian driving scenes.
An \emph{analytic tangent renderer} provides log brightness, its exact temporal derivative, and occlusion cues in a single pass.
An \emph{embedded Hermite controller} discards pixels that cannot reach a threshold within an interval, and bisects the remaining intervals until the crossings of a coarse cubic and two finer cubics agree within the tolerance.
\emph{Visibility-aware bracketing} isolates brightness discontinuities caused by occlusions, and \emph{sparse tile scheduling} re-renders only the tiles that require refinement while reusing cached samples.
Experiments on synthetic benchmarks, real recordings, and simulation-to-real optical flow demonstrate that \method achieves state-of-the-art timing accuracy while reducing rendering cost by one to two orders of magnitude.

The main contributions of this work are summarized as follows:
\begin{itemize}
    \item A reformulation of event simulation as error-controlled root finding, in which an embedded Hermite controller refines each interval until coarse and refined crossing estimates agree within a prescribed tolerance, making timing accuracy an explicit parameter.
    \item A visibility-aware bracketing strategy that localizes events at occlusion boundaries and moving objects instead of interpolating across or suppressing them.
    \item A sparse tile scheduling scheme with cached samples that restricts computation to regions where events can occur and scales to complete driving scenes and long sequences.
\end{itemize}

%% file: sec/2_related.tex
\section{Related Work}
\label{sec:related}

\paragraph{Event camera simulation.}
Most event simulators derive events from intensity frames.
ESIM~\cite{rebecq2018esim} couples a rendering engine with an adaptive sampling rate and linearly interpolates log brightness between samples, an event model also adopted by earlier simulators~\cite{mueggler2017event} and the CARLA event sensor~\cite{dosovitskiy2017carla}.
Vid2E~\cite{gehrig2020vid2e} applies this model to conventional videos after frame interpolation, while v2e~\cite{hu2021v2e}, ICNS~\cite{joubert2021icns}, and DVS-Voltmeter~\cite{lin2022dvsvoltmeter} add pixel bandwidth, noise, and stochastic circuit behavior to improve realism.
Learning-based methods~\cite{zhu2021eventgan,zhang2024v2ce} instead predict event representations from frames, and Stoffregen~\etal~\cite{stoffregen2020reducing} show that matching simulation parameters to real sensors reduces the sim-to-real gap.
These methods mainly improve the sensor model, whereas event times are still recovered from interpolated samples.
The resulting timing error depends on the frame rate or sampling interval and is not bounded for individual events, which motivates our error-controlled formulation.

\begin{figure*}[t]
  \centering
  \includegraphics[width=\linewidth]{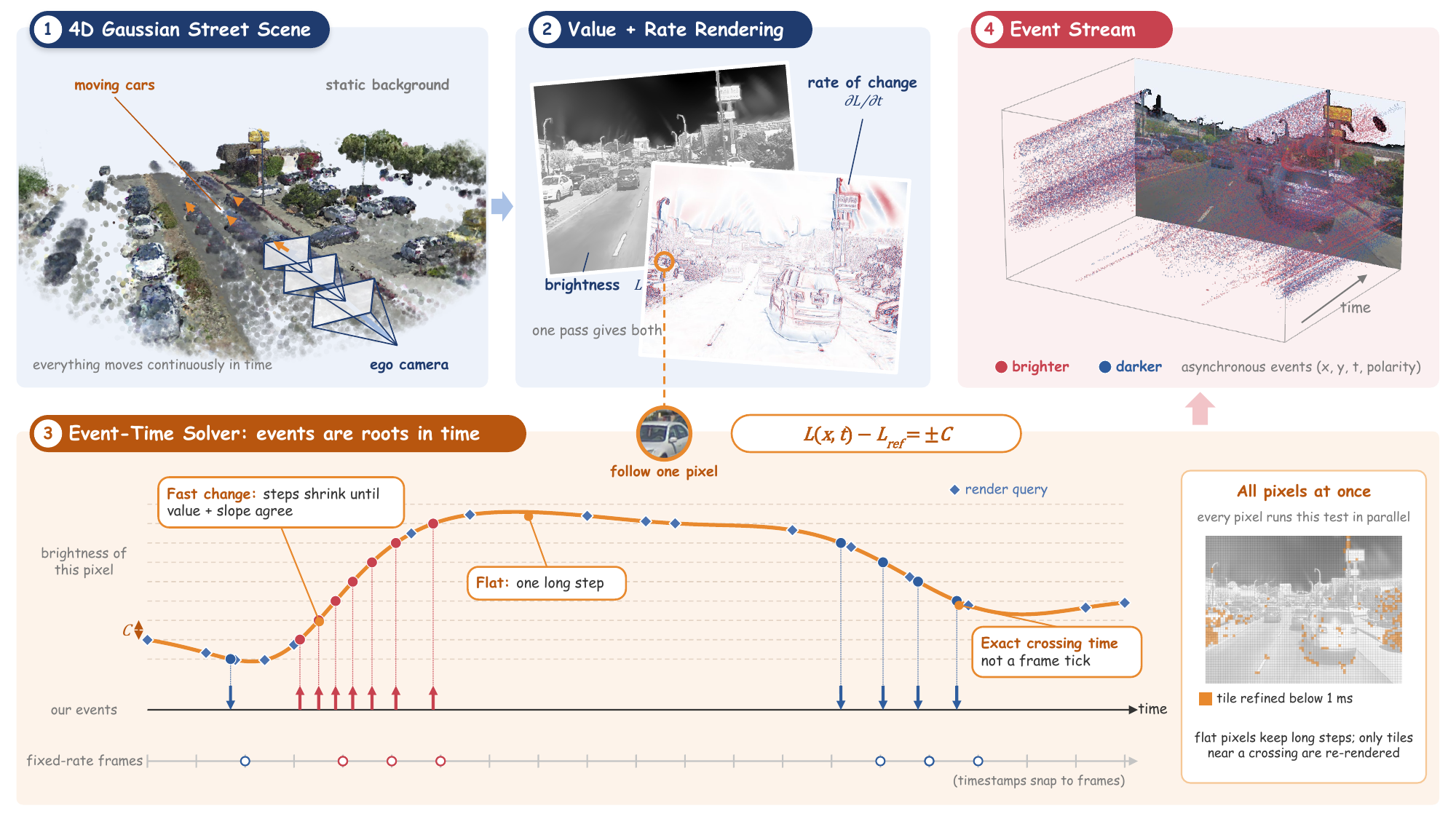}
  \caption{\textbf{Overview of \method.}
  (1)~A Gaussian driving scene evolves continuously in time.
  (2)~For any query time, the analytic tangent renderer returns the log brightness of each pixel and its temporal derivative in a single pass.
  (3)~The event-time solver treats each event as a root of the brightness threshold equation and refines the time step until value and slope agree, so events are placed at their crossing times rather than at frame ticks.
  (4)~All pixels are processed in parallel; pixels with slowly varying brightness keep long steps, and only tiles near a crossing are re-rendered.}
  \label{fig:overview}
\end{figure*}

\paragraph{Gaussian scene reconstruction and event rendering.}
Neural radiance fields~\cite{mildenhall2020nerf} and 3DGS~\cite{kerbl2023gaussian} enable photorealistic novel-view synthesis, and their dynamic extensions~\cite{yang2024deformable,wu20244dgs} model time-varying content.
For driving, scene graphs and dynamic decompositions~\cite{yang2023unisim,yang2024emernerf,tonderski2024neurad,yan2024streetgaussians,zhou2024drivinggaussian,zhou2024hugs,chen2023pvg,chen2025omnire} reconstruct static backgrounds and moving actors from sensor logs, which makes rendering at arbitrary times and poses possible.
Event cameras have also been used to supervise such representations~\cite{rudnev2023eventnerf,klenk2023enerf,hwang2023evnerf,xiong2024event3dgs,han2024event3dgs,yura2025eventsplat,feng2025e4dgs}, typically by comparing rendered brightness differences at two timestamps with accumulated events.
In the opposite direction, GS2E~\cite{li2025gs2e} renders dense frames from Gaussian scenes for a frame-driven event model.
TIDES~\cite{thirgood2026tides} obtains temporal derivatives by forward-mode differentiation, predicts possibly multiple crossings per step from a linear or quadratic model, shrinks its steps according to a risk score built from the visibility rate, contributor count, and coverage, and models sensor readout with a tile-level arbiter.
Its steps are thus chosen heuristically, and the timing error of the emitted events is neither estimated nor bounded by a user-specified tolerance.
\method shares the analytic derivatives but replaces risk-driven stepping by error-controlled root finding, localizes visibility changes within a bracket instead of only shortening steps around them, and renders only the tiles that require refinement.

\paragraph{Event location and adaptive sampling.}
Locating the time at which a solution crosses a threshold is a classical problem in numerical integration.
ODE solvers combine embedded error estimates~\cite{dormand1980family} with dense output to locate such events within a prescribed tolerance~\cite{hairer1993solving,shampine2000event}.
Rendering faces related problems: ray--surface intersection relies on interval bounds~\cite{kalra1989guaranteed,mitchell1990robust} or distance bounds~\cite{hart1996sphere} to exclude roots safely, adaptive sampling concentrates samples where the signal varies~\cite{mitchell1987generating,egan2009frequency}, and differentiable renderers treat visibility discontinuities explicitly~\cite{li2018differentiable,loubet2019reparameterizing}.
These techniques operate along rays or within individual integration steps and remain largely unexplored for the per-pixel brightness signal that defines events.
\method transfers them to event simulation: embedded estimates control timing error, brightness bounds exclude pixels and intervals without crossings, and visibility changes reported by the renderer are localized in time.

\begin{figure*}[t]
  \centering
  \includegraphics[width=\linewidth]{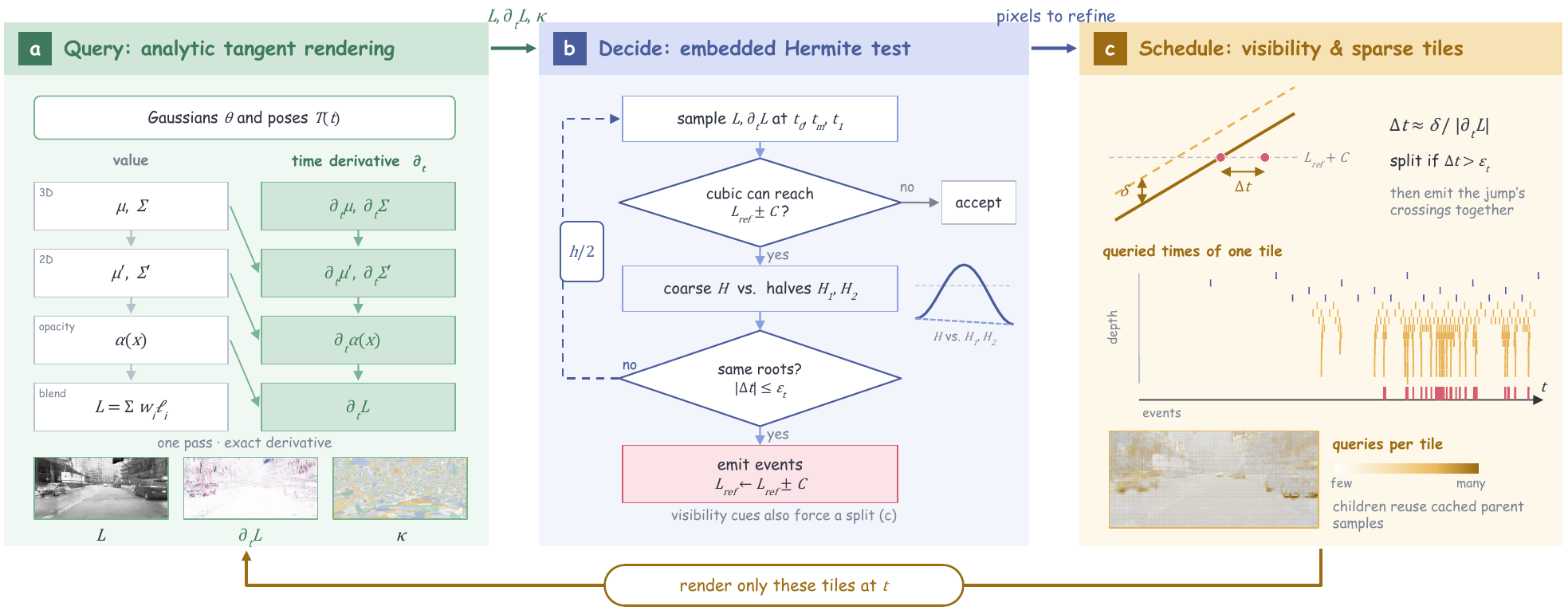}
  \caption{\textbf{Solver loop of \method.}
  (a)~\emph{Query}: forward-mode rendering propagates values and time derivatives through the 3D, 2D, opacity, and blending stages, yielding $L$, $\partial_t L$, and a visibility cue $\kappa$.
  (b)~\emph{Decide}: an interval is accepted when its cubic cannot reach a threshold level, or when the coarse and refined Hermite interpolants place the same crossings within $\varepsilon_t$; otherwise it is bisected.
  (c)~\emph{Schedule}: visibility changes force splits until the induced timing defect falls below $\varepsilon_t$, and each refinement renders only the requesting tiles while reusing cached parent samples.}
  \label{fig:solver}
\end{figure*}

%% file: sec/3_method.tex
\section{Method}
\label{sec:method}

\method simulates events by solving for the times at which the log brightness of each pixel crosses its contrast thresholds (\cref{sec:formulation}).
An analytic tangent renderer evaluates the log brightness, its exact temporal derivative, and a visibility cue at any query time (\cref{sec:tangent}).
These samples feed an embedded Hermite controller, which brackets every crossing by bisection until two interpolants of different accuracy agree on its time within a prescribed tolerance (\cref{sec:hermite}).
Occlusions are handled by visibility-aware bracketing, which converts brightness discontinuities into timing defects and localizes them in time (\cref{sec:visibility}).
Sparse tile scheduling executes the resulting per-pixel refinement on the few tiles that require it (\cref{sec:tiles}).
\Cref{fig:overview} illustrates the pipeline, and \cref{fig:solver} details the solver loop.

\subsection{Problem Formulation}
\label{sec:formulation}

The scene consists of a static background and rigid actors~\cite{yan2024streetgaussians,chen2025omnire} represented by Gaussians~\cite{kerbl2023gaussian}.
Actor Gaussians follow the interpolated pose of their actor, and the camera follows its own trajectory, so the scene can be rendered, together with its time derivatives, at any time $t$.
The log brightness of pixel $\mathbf{x}$ is $L(\mathbf{x},t)=\log(Y(\mathbf{x},t)+\epsilon)$, where $Y$ is the rendered luminance and $\epsilon$ a small constant.
An ideal event pixel stores a reference level $L_{\mathrm{ref}}(\mathbf{x})$ and fires an event of polarity $p\in\{+1,-1\}$ at the first time $\tau$ that satisfies
\begin{equation}
  L(\mathbf{x},\tau)-L_{\mathrm{ref}}(\mathbf{x})=p\,C,
  \label{eq:root}
\end{equation}
after which the reference is reset to $L_{\mathrm{ref}}+pC$.
The events of a pixel are thus the ordered roots of \cref{eq:root}, and the task is to locate every root within a time tolerance $\varepsilon_t$ using as few render queries as possible.
Comparator-level non-idealities such as noise, refractory periods, and threshold variation act on this ideal stream, whereas signal-level effects such as photoreceptor low-pass filtering modify $L$ and can be included before root finding.
Each pixel processes base intervals of length $\Delta$ in temporal order.

\subsection{Analytic Tangent Renderer}
\label{sec:tangent}

A render query at time $t$ returns $L$, its derivative $\dot L=\partial_t L$, and a visibility cue $\kappa$ (\cref{fig:solver}a).
As in TIDES~\cite{thirgood2026tides}, the time derivative is propagated in forward mode through every stage of Gaussian rasterization: pose derivatives give the velocities of 3D means and covariances, projection gives those of their 2D counterparts and of each opacity $\alpha_i$, and front-to-back compositing with transmittance $T_i$ gives
\begin{equation}
  \dot Y=\textstyle\sum_i \big(\dot y_i\,\alpha_i T_i+y_i\,(\dot\alpha_i T_i+\alpha_i\dot T_i)\big),
  \label{eq:blend}
\end{equation}
with per-Gaussian luminance $y_i$ and $\dot L=\dot Y/(Y+\epsilon)$ (full expressions in the supplement).
The tangent is accumulated in the same loop as the primal values, so a query costs about as much as an ordinary render.
\Cref{eq:blend} is exact while the set and order of contributing Gaussians stay fixed.
When they change, $L$ may jump or bend unpredictably, so the renderer also records in $\kappa$ the index of the dominant Gaussian, an order-sensitive hash of all contributing Gaussians, and their count; a change of $\kappa$ between samples signals a possible visibility event.
We further replace the hard opacity cut-off of rasterization by a smooth fade, so that jumps of $L$ stem from genuine visibility changes rather than from truncation (supplement).

\subsection{Embedded Hermite Controller}
\label{sec:hermite}

Consider an interval $[t_0,t_1]$ of length $h$ with midpoint $t_m$, where $L$ and $\dot L$ are sampled at all three times (\cref{fig:solver}b).
The coarse interpolant $H$ is the cubic Hermite polynomial matching both quantities at $t_0$ and $t_1$, and the refined interpolant $\tilde H$ consists of two such cubics on the halves.
Since cubic Hermite interpolation has an $\mathcal{O}(h^4)$ error, $\tilde H$ is markedly more accurate, and the deviation of $H$ at the midpoint,
\begin{equation}
  \delta=s\big(|L_m-H(t_m)|+\tfrac{h}{8}|\dot L_m-\dot H(t_m)|\big),
  \label{eq:defect}
\end{equation}
estimates the interpolation error with a safety factor $s$, mirroring embedded Runge--Kutta pairs~\cite{dormand1980family}.

\paragraph{Risk screen and agreement test.}
If neither the closed-form range of $H$ widened by $\delta$ nor the midpoint sample reaches a level $L_{\mathrm{ref}}\pm C$, the interval cannot produce an event and is accepted; most pixels leave the controller here.
Otherwise, the crossings $\tau_k$ of $H$ and $\tilde\tau_k$ of $\tilde H$ are computed in temporal order, and the interval is accepted if both agree in number and polarity and
\begin{equation}
  s\max_k|\tau_k-\tilde\tau_k|\le\varepsilon_t.
  \label{eq:agree}
\end{equation}
Accepted events are emitted at $\tilde\tau_k$, updating the reference after each event; otherwise the interval is bisected and its halves are processed in order.
The tolerance thus replaces the frame rate as the parameter that sets timing accuracy, and crossings with small slopes, whose timing is most sensitive, automatically receive deeper refinement.
Intervals that remain unresolved at a minimum length are reported rather than silently accepted.
Like local error estimates in adaptive ODE solvers, this criterion assumes that $L$ is resolved by its samples: a pulse that begins and ends between samples without affecting values, derivatives, or visibility cues cannot be detected, and we therefore validate the realized errors empirically (\cref{sec:exp_timing}).

\subsection{Visibility-Aware Bracketing}
\label{sec:visibility}

A visibility change inside an interval introduces a jump or kink of size $J$ into $L$.
Both interpolants smooth over it in a correlated manner, so they can agree with each other while both miss the true crossing by about $J/|\dot L|$; a step negligible in value can thus be large in time.
When $\kappa$ changes among the samples, the value defect is therefore converted to the time scale, and the interval is bisected before root finding whenever $h>b$ and
\begin{equation}
  \delta>\rho\,C\quad\text{or}\quad\delta/\sigma>\varepsilon_t,
  \label{eq:gate}
\end{equation}
where $\sigma$ is a lower estimate of the local slope from the samples, $b\le\varepsilon_t$ the bracket width, and $\rho$ a signal-scale ratio.
Recursion confines each visibility change to a bracket of width at most $b$, within which $L$ is interpolated linearly and all crossed levels are emitted in order; for a true jump, this defines the event time up to the bracket width.
Boundary events are thus localized within $b$ rather than interpolated across or suppressed.

\subsection{Sparse Tile Scheduling}
\label{sec:tiles}

The controller decides per pixel, whereas the renderer is efficient on tiles.
Each refinement round groups pending requests by time and renders only the tiles containing a requesting pixel (\cref{fig:solver}c).
Since base intervals start on the grid $t=k\Delta$ and are refined only by bisection, all query times lie on a dyadic grid shared by every pixel, which keeps rendering batches large.
A child interval reuses the cached samples at its parent's endpoints and midpoint, so each bisection costs one new query per child.
Beyond the base grid, computation thus scales with the number and difficulty of crossings rather than with a frame rate.

%% file: sec/4_experiments.tex
\section{Experiments}
\label{sec:experiments}

\begin{figure}[t]
  \centering
  \includegraphics[width=\linewidth]{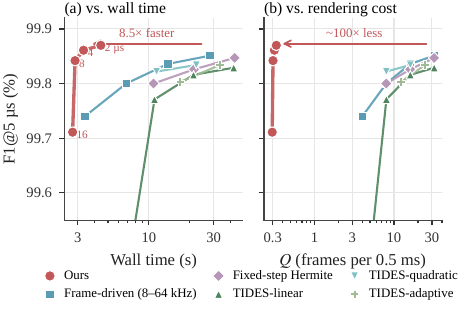}
  \caption{\textbf{Accuracy--cost frontier} on 8 ego-and-actor windows (exclusive GPU). \method varies only $\varepsilon_t$.}
  \label{fig:frontier}
\end{figure}

\begin{figure*}[t]
  \centering
  \includegraphics[width=\linewidth]{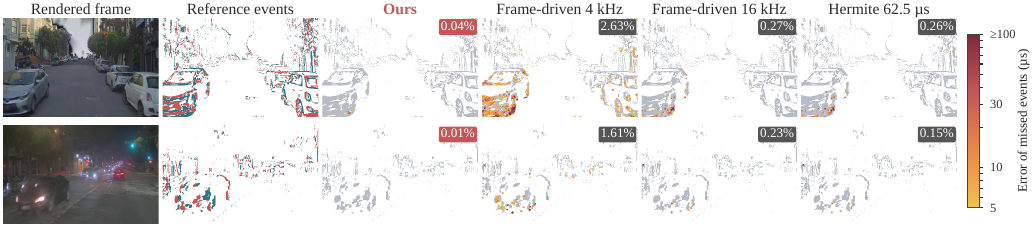}
  \caption{\textbf{Events missed at 5\,$\mu$s} (colored by timing error) over reference events (gray), for a day and a night window. Badges give miss rates.}
  \label{fig:qual_synthetic}
\end{figure*}

\subsection{Setup}
\label{sec:setup}

\paragraph{Data and metrics.}
All simulators run on the same Gaussian driving scenes reconstructed with OmniRe~\cite{chen2025omnire}, so that differences stem only from how event times are obtained.
Timing accuracy is measured on 8 PandaSet~\cite{xiao2021pandaset} scenes with 4 windows of 20\,ms and 3 motion types (ego, actor, both) at $346\times260$.
The reference is an independent dense rendering every 3.9\,$\mu$s, phase-shifted against all method grids, and events are matched per pixel and polarity within a time tolerance to compute F1.
Cost is the wall time and $Q$, the rendered area per 0.5\,ms in full frames ($Q=32$ for 64\,kHz frames).
Real-sensor generalization is evaluated on M3ED~\cite{chaney2023m3ed} at $1280\times720$ with the protocol and metrics of TIDES~\cite{thirgood2026tides}.

\paragraph{Baselines and details.}
We compare with frame-driven simulation that interpolates rendered frames~\cite{gehrig2020vid2e}, the official ESIM~\cite{rebecq2018esim}, v2e~\cite{hu2021v2e}, and DVS-Voltmeter~\cite{lin2022dvsvoltmeter}, fixed-step Hermite interpolation, and TIDES~\cite{thirgood2026tides}, which we re-implement on our renderer as its official code has not been released.
\method uses $\Delta=8$\,ms, $\varepsilon_t=4\,\mu$s, $s=2$, $b=2\,\mu$s, $\rho=0.025$, $8\times8$ tiles, and $C=0.2$.
All configurations were fixed before the experiments; further details and results are given in the supplement.

\begin{figure}[t]
  \centering
  \includegraphics[width=\linewidth]{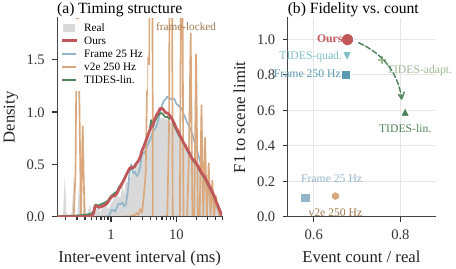}
  \caption{\textbf{M3ED analysis.} (a) Per-pixel inter-event intervals. (b) Fidelity to the scene limit versus event count.}
  \label{fig:real}
\end{figure}

\subsection{Event Timing Accuracy}
\label{sec:exp_timing}

\input{tables/tab_synthetic}

\paragraph{Accuracy at a fraction of the cost.}
As shown in \cref{tab:synthetic}, \method reaches 99.88\% F1 at 5\,$\mu$s and 99.97\% at 10\,$\mu$s, on par with 64\,kHz frames and 31.25\,$\mu$s Hermite interpolation, while rendering about 1/100 and 1/50 of their image area and running 5$\times$ and 4$\times$ faster.
Frame-driven simulation trades accuracy for cost (\cref{fig:frontier}): at 4\,kHz, its miss rate grows by more than an order of magnitude.
v2e places events at frame times and stays below 3\% F1.
TIDES is evaluated without its opacity-based emission gate, which otherwise suppresses faint actor edges and limits its F1 to 87\%; even so, its most accurate variant needs 6.4$\times$ the time and 74$\times$ the rendering cost of \method.

\begin{figure*}[t]
  \centering
  \includegraphics[width=\linewidth]{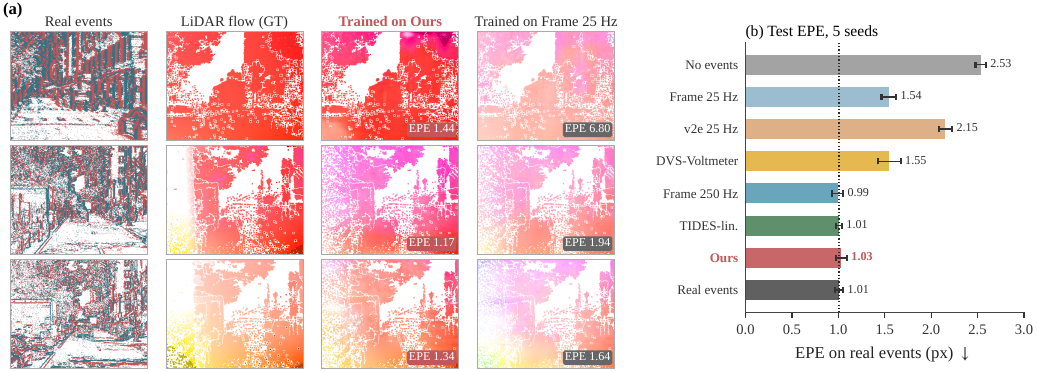}
  \caption{\textbf{Sim-to-real optical flow on real M3ED events.} (a) Flow of networks trained on simulated events, shown on the LiDAR support, with per-sample EPE. (b) Test EPE over 5 seeds; the dotted line marks training on real events.}
  \label{fig:flow}
\end{figure*}

\paragraph{Why error control matters.}
The gap to TIDES-adaptive stems from what drives refinement.
Its risk score reacts to visibility changes and depth competition whether or not a crossing occurs, and a single step size serves all pixels, so refinement spreads over the whole image and the cost reaches $Q=23$.
\method refines a pixel only when its own crossing estimates disagree or a visibility change can shift a crossing by more than $\varepsilon_t$, which yields slightly higher accuracy with 74$\times$ less rendering.

\paragraph{Frontier.}
On the 8 frontier windows with an exclusive GPU (\cref{fig:frontier}), \method with $\varepsilon_t=4\,\mu$s attains the highest F1 of all configurations (99.861\% versus 99.851\% for 64\,kHz frames) at 8.5$\times$ less time and about 100$\times$ less rendering; absolute values differ from \cref{tab:synthetic}, which pools all 96 windows.
The single parameter $\varepsilon_t$ traces the entire trade-off, and all baselines lie below this frontier.

\paragraph{Calibration.}
Re-simulating four frontier windows against a reference with a four times finer step shows that the tolerance controls the realized error: at $\varepsilon_t=4\,\mu$s, only 0.27\% of the events deviate by more than $\varepsilon_t$, and this fraction decreases monotonically with $\varepsilon_t$ (supplement).
The standard reference agrees with the finer one within 1\,$\mu$s for 99\% of the events, and against the finer reference \method still misses fewer events than 64\,kHz frames.

\paragraph{Qualitative results.}
\Cref{fig:qual_synthetic} shows that the failures of frame-driven simulation concentrate on vehicle boundaries and fine textures, where brightness changes faster than the frame interval.
\method misses 0.04\% and 0.01\% of the events in these windows, 7--18$\times$ fewer than 16\,kHz frames.
The parity with dense sampling also holds per window: in 70\% of the windows, the F1 of \method lies within $\pm0.05$ percentage points of 64\,kHz frames at a median speed-up of 5.7$\times$ (supplement).

\input{tables/tab_ablation}

\subsection{Ablation Study}
\label{sec:ablation}

\Cref{tab:ablation} shows that each component has a distinct effect.
Without the embedded controller, fixed-step sampling needs 25$\times$ the rendering cost for similar accuracy, and dropping the tangent as well further increases misses by 25\%.
Without visibility-aware bracketing, misses grow by 4.2$\times$, and the timing gate of \cref{eq:gate} alone accounts for most of this gain, confirming that visibility steps negligible in value matter in time.
Sparse tile scheduling reduces the cost by 21$\times$ without changing a single event.

\begin{figure*}[t]
  \centering
  \includegraphics[width=\linewidth]{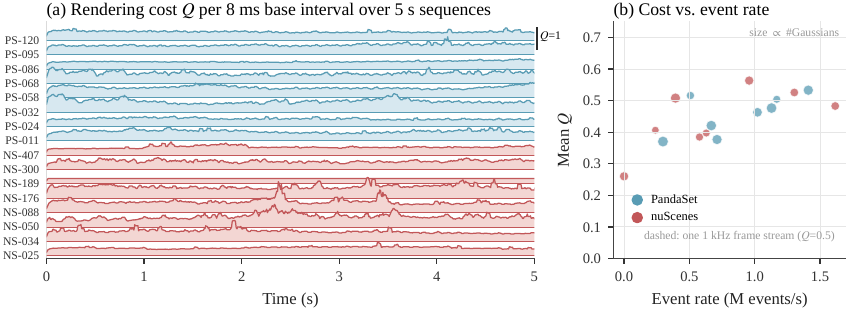}
  \caption{\textbf{Long sequences.} (a) Rendering cost per 8\,ms base interval over 5\,s for 16 complete PandaSet (PS) and nuScenes (NS) scenes. (b) Mean cost versus event rate; marker size indicates the number of Gaussians.}
  \label{fig:long}
\end{figure*}

\subsection{Generalization to Real Recordings}
\label{sec:exp_real}

\input{tables/tab_m3ed}

\Cref{tab:m3ed,fig:real} evaluate 24 windows of 50\,ms from the \emph{horse} and \emph{rittenhouse} sequences with 3 phase seeds; each scene is reconstructed without data within $\pm100$\,ms of its window.
\method obtains the best mean rank over six real-data metrics (\cref{tab:m3ed}), and simulation at the 25\,Hz camera rate falls clearly behind.
\Cref{fig:real}a shows that \method follows the real inter-event-interval distribution, whereas v2e locks timestamps to the frame grid.
TIDES-linear leads on count-based metrics, but \cref{fig:real}b explains why: its extrapolation overshoots, adding 5.6\% events and reaching only 0.59 F1 to the scene's own continuous-time limit, versus 0.998 for \method.
As reconstructed scenes yield about 68\% of the real event count, such surplus events mimic sensor effects, so real-data metrics should be read together with scene fidelity.
Since real data do not need microsecond accuracy, relaxing $\varepsilon_t$ to 250\,$\mu$s cuts the time by 5.5$\times$ with unchanged metrics.

\subsection{Sim-to-Real Optical Flow}
\label{sec:exp_flow}

A U-Net~\cite{ronneberger2015unet} regresses optical flow from two 50\,ms event voxel grids, and each simulator is trained with 5 seeds.
It is trained on events simulated from the M3ED \emph{penno} and \emph{horse} scenes with flow from rendered depth and poses, and tested on real \emph{rittenhouse} events with LiDAR-based flow; the protocol was accepted only after training on real events reduced the end-point error (EPE) by more than 20\% over an input without events.
Training on \method events yields an EPE of $1.03\pm0.06$\,px, matching training on real events ($1.01\pm0.04$\,px), whereas video-rate simulation and DVS-Voltmeter reach 1.54--2.15\,px and an input without events 2.53\,px (\cref{fig:flow}b).
Continuous-time simulators perform on par, which indicates that flow at this scale benefits from continuous timing rather than from microsecond accuracy.
\Cref{fig:flow}a shows that networks trained on video-rate simulation underestimate fast ego motion, while training on \method recovers the magnitude and the depth-dependent structure of the ground truth.

\subsection{Scalability to Long Sequences}
\label{sec:exp_long}

\method simulates 16 complete scenes from PandaSet and nuScenes~\cite{caesar2020nuscenes} with up to 2.2 million Gaussians, each for 0.5\,s and 5\,s.
All 32 runs finish without unresolved intervals, produce up to 8 million events, and use at most 10.3\,GB of GPU memory.
As shown in \cref{fig:long}a, the cost stays stable over time and rises only briefly when large objects cross the view.
Its mean lies between $Q=0.26$ and 0.66, comparable to a single 1\,kHz frame stream, and grows with the event rate rather than with the duration or the number of Gaussians (\cref{fig:long}b).
Since refinement follows the crossings, long and complex scenes are simulated at microsecond accuracy without a frame-rate budget.

%% file: tables/tab_synthetic.tex
\begin{table}[t]
\centering
\caption{\textbf{Timing accuracy on synthetic scenes} ($346\times260$, 96 windows). F1 (\%) at 5 and 10\,$\mu$s; time (s) per 20\,ms window. $^\dagger$Without emission gate. \colorbox{rkA}{\textbf{1st}} \colorbox{rkB}{2nd} \colorbox{rkC}{3rd}.}
\label{tab:synthetic}
\footnotesize
\setlength{\tabcolsep}{3.4pt}
\begin{tabular}{llcccc}
\toprule
Method & Rate & F1@5$\uparrow$ & F1@10$\uparrow$ & Time$\downarrow$ & $Q\downarrow$\\
\midrule
Frame~\cite{gehrig2020vid2e} & 4\,kHz & 97.707 & 99.206 & \fst{1.19} & \snd{2.00} \\
Frame~\cite{gehrig2020vid2e} & 16\,kHz & 99.850 & 99.942 & \trd{4.53} & \trd{8.00} \\
Frame~\cite{gehrig2020vid2e} & 64\,kHz & \fst{99.908} & \fst{99.979} & 18.07 & 32.00 \\
ESIM~\cite{rebecq2018esim} & 4\,kHz & 97.708 & 99.207 & 9.64 & \snd{2.00} \\
v2e-HDR~\cite{hu2021v2e} & 4\,kHz & 2.541 & 5.102 & 7.27 & \snd{2.00} \\
Hermite & 62.5\,$\mu$s & 99.871 & 99.950 & 6.53 & \trd{8.00} \\
Hermite & 31.25\,$\mu$s & \snd{99.889} & \trd{99.967} & 13.54 & 16.00 \\
TIDES-lin.$^\dagger$~\cite{thirgood2026tides} & 16\,kHz & 99.713 & 99.879 & 6.96 & 8.03 \\
TIDES-quad.$^\dagger$~\cite{thirgood2026tides} & 16\,kHz & 99.869 & 99.949 & 7.37 & 8.03 \\
TIDES-adapt.$^\dagger$~\cite{thirgood2026tides} & adaptive & 99.875 & 99.953 & 22.66 & 23.24 \\
\midrule
\textbf{\method (Ours)} & $\varepsilon_t$=4\,$\mu$s & \trd{99.881} & \snd{99.973} & \snd{3.55} & \fst{0.32} \\
\bottomrule
\end{tabular}
\end{table}

%% file: tables/tab_ablation.tex
\begin{table}[t]
\centering
\caption{\textbf{Ablation} at $346\times260$. Each block removes one component; degradations relative to the full model are given in red. $^\dagger$Main benchmark; other rows on held-out scenes.}
\label{tab:ablation}
\footnotesize
\setlength{\tabcolsep}{4.5pt}
\renewcommand{\arraystretch}{1.1}
\begin{tabular}{cl ll}
\toprule
& Variant & Miss@5\,$\mu$s (\%)$\downarrow$ & $Q\downarrow$\\
\midrule
\rowcolor{rkC} & \textbf{\method (full)} & \textbf{0.12} & \textbf{0.33}\\
\midrule
\multirow{2}{*}{\rotatebox{90}{\scriptsize Root}} & w/o controller$^\dagger$ & 0.13 & 8.00\,\dg{25}\\
& w/o controller \& tangent$^\dagger$ & 0.15\,\dg{1.3} & 8.00\,\dg{25}\\
\midrule
\multirow{2}{*}{\rotatebox{90}{\scriptsize Vis.}} & w/o bracketing & 0.51\,\dg{4.2} & 0.31\\
& w/o timing gate (\cref{eq:gate}) & 0.43\,\dg{3.6} & 0.31\\
\midrule
\multirow{2}{*}{\rotatebox{90}{\scriptsize Tiles}} & w/o sparse scheduling & 0.12 & 7.03\,\dg{21}\\
& $16\times16$ tiles & 0.12 & 0.44\,\dg{1.3}\\
\bottomrule
\end{tabular}
\end{table}

%% file: tables/tab_m3ed.tex
\begin{table}[t]
\centering
\caption{\textbf{Real recordings} (M3ED, 24 windows $\times$ 3 seeds). Rank: mean rank over six real-data metrics (full table in the supplement). Scene: F1@0.25\,ms to the scene's continuous-time limit.}
\label{tab:m3ed}
\footnotesize
\setlength{\tabcolsep}{2.6pt}
\begin{tabular}{lcccccc}
\toprule
Method & Voxel$\uparrow$ & T-surf.$\uparrow$ & Count$\downarrow$ & $W_1\downarrow$ & Rank$\downarrow$ & Scene$\uparrow$\\
\midrule
Frame 25\,Hz & 0.1110 & 0.3160 & 0.466 & 0.507 & 8.33 & 0.104 \\
Frame 250\,Hz & \fst{0.1322} & 0.3291 & 0.432 & 0.403 & 4.67 & 0.798 \\
v2e-HDR 250\,Hz & 0.1266 & 0.3277 & 0.452 & 0.407 & 6.00 & 0.117 \\
DVS-Voltmeter & 0.0926 & 0.2288 & 0.918 & \fst{0.376} & 5.80 & 0.014 \\
TIDES-lin. & 0.1297 & \fst{0.3411} & \fst{0.378} & 0.425 & \snd{3.67} & 0.591 \\
TIDES-quad. & \trd{0.1320} & 0.3311 & 0.423 & \snd{0.390} & \trd{3.83} & \snd{0.904} \\
TIDES-adapt. & 0.1313 & \snd{0.3370} & \snd{0.393} & 0.397 & \snd{3.67} & \trd{0.884} \\
\midrule
\textbf{\method (Ours)} & \snd{0.1321} & \trd{0.3319} & \trd{0.416} & \trd{0.391} & \fst{3.00} & \fst{0.998} \\
\bottomrule
\end{tabular}
\end{table}

%% file: sec/5_conclusion.tex
\section{Conclusion}
\label{sec:conclusion}

This paper presented \method, a continuous-time event simulator that solves event times as roots of the brightness threshold equation under explicit error control.
Its tangent renderer, embedded Hermite controller, visibility-aware bracketing, and sparse tile scheduling make the time tolerance the single parameter for accuracy and cost.
\method achieves state-of-the-art timing accuracy with one to two orders of magnitude less rendering on synthetic driving scenes, generalizes to real recordings, and supports sim-to-real optical flow on par with real data.

Still, \method models an ideal sensor on reconstructed scenes, which are smoother than the real world, and its error estimate, although well calibrated in our experiments, is empirical rather than a strict bound.
Calibrated sensor noise models and conservative visibility bounds are promising directions to further narrow the gap to real sensors.

%% file: sec/6_llm.tex
\section*{LLM Usage Statement}
Large language models were used to assist with language editing and drafting of the text. All methods, experiments, results, and claims were designed, verified, and are taken responsibility for by the authors.

%% file: sec/X_suppl.tex
\clearpage
\setcounter{page}{1}
\maketitlesupplementary
\renewcommand{\topfraction}{0.9}\renewcommand{\dbltopfraction}{0.9}\renewcommand{\bottomfraction}{0.6}
\renewcommand{\textfraction}{0.07}\renewcommand{\floatpagefraction}{0.85}\renewcommand{\dblfloatpagefraction}{0.85}
\setcounter{topnumber}{3}\setcounter{dbltopnumber}{2}\setcounter{totalnumber}{5}

\renewcommand{\thesection}{\Alph{section}}
\setcounter{section}{0}
\renewcommand{\thefigure}{S\arabic{figure}}
\setcounter{figure}{0}
\renewcommand{\thetable}{S\arabic{table}}
\setcounter{table}{0}
\renewcommand{\theequation}{S\arabic{equation}}
\setcounter{equation}{0}

This supplement provides the details omitted from the main paper.
\Cref{sec:supp_scene,sec:supp_tangent,sec:supp_controller,sec:supp_visibility,sec:supp_schedule} present the scene model and each component of \method, \cref{sec:supp_impl} the implementation, \cref{sec:supp_synthetic,sec:supp_real,sec:supp_flow,sec:supp_long} the protocols and additional results of each experiment, and \cref{sec:supp_discussion} further discussion of limitations.
Besides tables, the supplement contains visual results that complement the main paper from several angles:
event streams of all eight synthetic scenes (\cref{fig:supp_gallery}) and of the three motion types (\cref{fig:supp_motion});
the distribution of timing errors over 1.4\,M events (\cref{fig:supp_timing});
single pixels that show how each failure mode arises and how \method avoids it (\cref{fig:supp_pixels});
where the solver spends its renders (\cref{fig:supp_queries});
spatial maps of the events lost by each ablated component (\cref{fig:supp_ablation_maps});
space--time slices and event rates on real recordings (\cref{fig:supp_m3ed_xt});
additional optical-flow predictions together with a breakdown by motion magnitude (\cref{fig:supp_flow});
and snapshots of complete 5\,s sequences (\cref{fig:supp_long}).
None of these figures repeats a figure of the main paper.

\section{Scene and Signal Model}
\label{sec:supp_scene}

\paragraph{Gaussian driving scenes.}
Synthetic and long-sequence experiments use OmniRe~\cite{chen2025omnire} reconstructions of PandaSet~\cite{xiao2021pandaset} and nuScenes~\cite{caesar2020nuscenes}, which decompose a scene into a static background and rigid actors.
Each Gaussian $i$ has a canonical mean $\boldsymbol{\mu}_i^0$, covariance $\boldsymbol{\Sigma}_i^0$, opacity $o_i$, and spherical-harmonic color.
A Gaussian attached to actor $a$ is transformed by the actor pose $(\mathbf{R}_a(t),\mathbf{p}_a(t))$,
\begin{equation}
  \boldsymbol{\mu}_i(t)=\mathbf{R}_a(t)\boldsymbol{\mu}_i^0+\mathbf{p}_a(t),\qquad
  \boldsymbol{\Sigma}_i(t)=\mathbf{R}_a(t)\boldsymbol{\Sigma}_i^0\mathbf{R}_a(t)^{\top},
\end{equation}
whereas background Gaussians are static.
Actor and camera poses are given at keyframes and interpolated with spherical linear interpolation for rotations and linear interpolation for translations, so that poses and their time derivatives are available in closed form at any time.
All poses are evaluated in double precision, which avoids the loss of sub-microsecond resolution that occurs when absolute timestamps are stored in single precision.

\paragraph{Log brightness.}
The color of each Gaussian is converted to luminance $y_i=0.299\,r_i+0.587\,g_i+0.114\,b_i$ before compositing, so that the renderer composites a single channel.
The log brightness is $L=\log(Y+\epsilon)$ with $\epsilon=10^{-6}$.
On M3ED, a radiometric response is applied to $Y$ before the logarithm; its form and the contrast threshold are calibrated on windows disjoint from all test windows (\cref{sec:supp_m3ed_protocol}).

\paragraph{Ideal sensor.}
Each pixel stores a reference level and emits an event whenever $L$ departs from it by the contrast threshold $C$ (Eq.~(1) of the main paper), after which the reference moves by $\pm C$.
Several events may occur within one interval and are emitted in temporal order with successive references.
We use symmetric thresholds $C_+=C_-=C$; asymmetric thresholds only change the set of levels and are supported without modification.

\section{Analytic Tangent Renderer}
\label{sec:supp_tangent}

\paragraph{Projection.}
Let $\mathbf{W}(t)$ denote the world-to-camera transform and $\mathbf{J}(t)$ the Jacobian of the perspective projection at the camera-space mean.
The 2D mean $\boldsymbol{\mu}'_i$ is the projection of $\mathbf{W}\boldsymbol{\mu}_i$, and the 2D covariance is $\boldsymbol{\Sigma}'_i=\mathbf{J}\mathbf{W}\boldsymbol{\Sigma}_i\mathbf{W}^{\top}\mathbf{J}^{\top}$.
Their time derivatives follow from the product rule with $\dot{\boldsymbol{\mu}}_i=\dot{\mathbf{R}}_a\boldsymbol{\mu}_i^0+\dot{\mathbf{p}}_a$, $\dot{\boldsymbol{\Sigma}}_i=\dot{\mathbf{R}}_a\boldsymbol{\Sigma}_i^0\mathbf{R}_a^{\top}+\mathbf{R}_a\boldsymbol{\Sigma}_i^0\dot{\mathbf{R}}_a^{\top}$, the camera velocity $\dot{\mathbf{W}}$, and the derivative of $\mathbf{J}$ with respect to the camera-space mean.
View-dependent colors are differentiated through the spherical-harmonic basis, which gives $\dot y_i$.
All of these quantities are evaluated with forward-mode dual numbers in the same projection kernel that computes the primal values.

\paragraph{Opacity and compositing.}
With $\mathbf{d}_i=\mathbf{x}-\boldsymbol{\mu}'_i$ and $\mathbf{A}_i={\boldsymbol{\Sigma}'_i}^{-1}$, the opacity at pixel $\mathbf{x}$ is $\alpha_i=o_i\exp(-q_i)$ with $q_i=\tfrac12\mathbf{d}_i^{\top}\mathbf{A}_i\mathbf{d}_i$, and
\begin{equation}
  \dot\alpha_i=\dot o_i e^{-q_i}-\alpha_i\dot q_i,\qquad
  \dot q_i=-\mathbf{d}_i^{\top}\mathbf{A}_i\dot{\boldsymbol{\mu}}'_i-\tfrac12\mathbf{d}_i^{\top}\mathbf{A}_i\dot{\boldsymbol{\Sigma}}'_i\mathbf{A}_i\mathbf{d}_i .
\end{equation}
With weights $w_i=\alpha_iT_i$, front-to-back compositing and its tangent read
\begin{align}
  Y&=\textstyle\sum_i y_i w_i + T_N\,y_{\mathrm{bg}},\\
  \dot Y&=\textstyle\sum_i (\dot y_i w_i+y_i\dot w_i)+\dot T_N\,y_{\mathrm{bg}},\\
  T_{i+1}&=T_i(1-\alpha_i),\quad \dot T_{i+1}=\dot T_i(1-\alpha_i)-T_i\dot\alpha_i,
\end{align}
with $\dot w_i=\dot\alpha_iT_i+\alpha_i\dot T_i$ and background luminance $y_{\mathrm{bg}}$.
Both recursions run in one loop over the depth-sorted Gaussians of a tile, so the tangent adds a constant number of operations per Gaussian and no additional memory traffic.

\paragraph{Smooth support.}
Standard Gaussian rasterization discards contributions with $\alpha_i<\alpha_{\min}$ and terminates when the transmittance becomes negligible.
Both rules make $Y$ discontinuous in time, as a Gaussian crosses the cut-off.
We replace the hard opacity cut-off by a $C^1$ fade, $\alpha_i\leftarrow\alpha_i\,\phi\big((\alpha_i-\alpha_{\min})/\alpha_{\min}\big)$ for $\alpha_i<2\alpha_{\min}$ with the smoothstep $\phi(u)=u^2(3-2u)$, and the transmittance cut-off by an analogous smooth tail, both with threshold $10^{-5}$.
The modification changes rendered values by less than $10^{-5}$ but removes artificial discontinuities, so that the remaining jumps of $L$ stem from genuine visibility changes.

\paragraph{Visibility cue.}
During compositing, the renderer records for every pixel the index of the Gaussian with the largest weight $w_i$, an order-sensitive hash of the indices of all Gaussians with positive weight, and the number of Gaussians with $w_i>10^{-8}$.
An interval carries a visibility cue if any of these three quantities differs among its endpoint and midpoint samples.
Two signal-based tests complement this cue: the accumulated opacity at the midpoint deviates from the endpoint average by more than 0.05, or the midpoint value violates the trapezoidal rule on either half, $|L_m-L_0-\tfrac h4(\dot L_0+\dot L_m)|>0.1C$ or the analogous condition on the right half.
Because the hash depends on the order of contributors, a change of the depth order is detected even when the dominant Gaussian and the count stay the same.
The hash is a risk cue rather than a certificate, so a collision can only delay, not prevent, refinement triggered by the other tests.

\section{Embedded Hermite Controller}
\label{sec:supp_controller}

\paragraph{Interpolants.}
On an interval $[t_0,t_1]$ of length $h$, the coarse interpolant is
\begin{equation}
  H(t_0+uh)=h_{00}(u)L_0+h_{10}(u)\,h\dot L_0+h_{01}(u)L_1+h_{11}(u)\,h\dot L_1,
\end{equation}
with $h_{00}=2u^3-3u^2+1$, $h_{10}=u^3-2u^2+u$, $h_{01}=-2u^3+3u^2$, and $h_{11}=u^3-u^2$.
The refined interpolant $\tilde H$ applies the same formula on $[t_0,t_m]$ and $[t_m,t_1]$.
For a smooth signal, the interpolation error of $H$ is bounded by $\tfrac{h^4}{384}\max|L^{(4)}|$, so halving the interval reduces it by a factor of about 16.
The value and slope deviations of $H$ at the midpoint therefore estimate the error of $H$ and, conservatively, that of $\tilde H$; the factor $h/8$ in Eq.~(3) converts the slope deviation into the value deviation it induces over a quarter interval.

\paragraph{Risk screen.}
The extrema of the cubic $H$ on $[0,1]$ are obtained from the roots of its quadratic derivative, which gives its range $[H_{\min},H_{\max}]$ in closed form.
An interval is considered risky if $H_{\max}+\delta\ge L_{\mathrm{ref}}+C$, $H_{\min}-\delta\le L_{\mathrm{ref}}-C$, or the midpoint sample itself reaches a level; a margin of $2\times10^{-6}$ absorbs floating-point round-off.
Intervals that are not risky are accepted without root finding and leave the reference unchanged.

\paragraph{Crossings and agreement.}
For a risky interval, the crossings of $H$ and of both halves of $\tilde H$ with the successive levels $L_{\mathrm{ref}}\pm C$ are computed in temporal order.
Each crossing is a root of a cubic in $u$; all real roots in $[0,1]$ are computed in double precision.
After each crossing, the reference moves by $\pm C$ and the search continues from the crossing time, so that several events per interval are handled exactly.
The two event lists are compared per pixel: they must contain the same number of events with the same polarities, and corresponding events must satisfy Eq.~(4) of the main paper.
If any condition fails, the interval is bisected.
Accepted events are taken from $\tilde H$, which is the more accurate interpolant.
The safety factor $s=2$ follows common practice for embedded error estimates; the tolerance sweep in \cref{sec:supp_calibration} shows how the realized error changes with the tolerance.

\paragraph{Termination.}
Bisection stops at a minimum interval of $\Delta/2^{16}\approx0.12\,\mu$s, with a maximum of 17 levels.
An interval that fails the agreement test at this length is first re-evaluated with double-precision rendering.
If it still fails, the crossings of $\tilde H$ are accepted because their times are confined to an interval far shorter than $\varepsilon_t$, and the pixel is recorded; a run aborts if more than 64 pixels require this rule.
In the long sequences, this rule was invoked for a single pixel in each of PandaSet scenes 011 and 032.

\paragraph{Optional safeguards.}
Two further safeguards are enabled by default but did not change any result (\cref{sec:supp_ablation}).
First, intervals that appear flat, with a sample range below $0.008\,C$, but contain potential contributors that are not yet visible, are probed at their quarter points to detect pulses that could hide between samples.
Second, if the estimated single-precision noise of $L$ divided by the local slope exceeds $\varepsilon_t$, the interval is evaluated in double precision before acceptance.

\paragraph{Root solver on the GPU.}
Each pixel is handled by one CUDA thread in double precision.
The thread first splits the cubic at the roots of its quadratic derivative, so that it is monotone on each of at most three pieces, and then searches the pieces in temporal order.
On a piece that brackets the next level, the crossing is found by bisection until the bracket is shorter than $10^{-11}$\,s; we use bisection rather than Newton's method because it cannot leave the bracket and its termination is defined in time, not in the residual of $L$, which is uninformative where the slope is small.
A crossing whose derivative has the wrong sign, \ie a tangential touch, is not emitted.
The coarse cubic and the two refined cubics are traversed by three cursors in lockstep, so that the $j$-th crossing of $H$ is compared with the $j$-th crossing of $\tilde H$ without storing intermediate event lists.
Output is written in two passes: the first counts the events of each pixel, a prefix sum allocates the output, and the second writes the events, so that no crossing can be dropped because of a fixed buffer size.
Acceptance, the commit of the reference levels, and the compaction of the pixels to be bisected are performed in the same kernel, and the host reads back a single packet of counters per refinement level.

\paragraph{Algorithm.}
For every base interval $[k\Delta,(k+1)\Delta]$, the solver proceeds as follows.
(i)~Render the endpoints for all pixels and the midpoint for all active pixels.
(ii)~Compute $\delta$, the risk screen, and the visibility cue for each pixel.
(iii)~Accept non-risky pixels; bisect pixels that trigger the visibility gate; emit bracketed events for visible changes confined to width $b$.
(iv)~Solve the crossings of $H$ and $\tilde H$ for the remaining pixels and accept those that agree.
(v)~Recurse on the left and then the right half for all rejected pixels, reusing cached samples.
The reference level of each pixel is committed only after its interval has been accepted, so rejected trial events never enter the output.
\Cref{alg:supp_solver} summarizes the procedure.

\begin{algorithm}[t]
\caption{\method on one base interval}
\label{alg:supp_solver}
\footnotesize
\begin{algorithmic}[1]
\Require $I_0=[t_0,t_0+\Delta]$, references $L_{\mathrm{ref}}$, $\varepsilon_t$, $s$, $\rho$, $b$
\State $\mathcal{S}\gets\{(I_0,\text{all pixels})\}$ \Comment{stack of (interval, pixels)}
\While{$\mathcal{S}\neq\emptyset$}
  \State pop $([t_a,t_b],P)$; \ $t_m\gets\frac{t_a+t_b}{2}$; \ $h\gets t_b-t_a$
  \State render $L,\dot L$ at $t_a,t_m,t_b$ on the tiles of $P$ \Comment{cached}
  \ForAll{$p\in P$ \textbf{in parallel}}
    \State $\delta\gets$ main Eq.~(3); \ $\sigma\gets$ \cref{eq:supp_sigma}
    \If{not risky} accept $p$ without event
    \ElsIf{cue $\wedge$ $h\le b$ $\wedge$ monotone} bracketed emission
    \ElsIf{cue $\wedge$ $(\delta/\sigma>\varepsilon_t \vee \delta>\rho C)$} reject $p$
    \Else
      \State $E\gets$ crossings of $H$; \ $\tilde E\gets$ crossings of $\tilde H$
      \If{$E,\tilde E$ agree (main Eq.~(4))} emit $\tilde E$; update $L_{\mathrm{ref}}(p)$
      \Else{} reject $p$
      \EndIf
    \EndIf
  \EndFor
  \State $R\gets$ rejected pixels
  \If{$R\neq\emptyset$ $\wedge$ $h/2\ge h_{\min}$} push right, then left half with $R$
  \ElsIf{$R\neq\emptyset$} fp64 re-check, else accept $\tilde E$ and log $p$
  \EndIf
\EndWhile
\end{algorithmic}
\end{algorithm}

\paragraph{Ordering and state.}
Because the left half is always resolved before the right half, the reference levels that enter the right half are those after all events of the left half, and the output of each pixel is in temporal order without sorting.
Only the reference level and the samples at the end of the base interval are carried to the next interval, so the state per pixel is constant and independent of the number of events.

\section{Visibility-Aware Bracketing}
\label{sec:supp_visibility}

\paragraph{Why a time-scale gate is needed.}
Consider a weak contributor that enters a pixel and changes $L$ by a step $J$ at time $t_J$.
If the step is small compared with the curvature of $L$, both $H$ and $\tilde H$ absorb it into a smooth cubic, and their crossings may agree to within less than 1\,$\mu$s while both are displaced from the true crossing by about $J/|\dot L|$.
In our diagnostics, steps of $10^{-4}$ to $10^{-3}$ at slopes of about $30\,\mathrm{s}^{-1}$ displaced crossings by 5--20\,$\mu$s, whereas a gate on the value scale alone, $\delta>\rho C$ with $\rho C=5\times10^{-3}$, did not trigger.

\paragraph{Timing defect.}
The slope in Eq.~(5) is estimated as
\begin{equation}
  \label{eq:supp_sigma}
  \sigma=\max\Big(\frac{|L_1-L_0|}{h},\ \min\big(|\dot L_0|,|\dot L_m|,|\dot L_1|\big)\Big),
\end{equation}
which is a lower estimate of $|\dot L|$ on the interval and thus an upper estimate of the timing defect $\delta/\sigma$.
The gate is applied only when the interval carries a visibility cue and is risky, so smooth intervals keep their long steps.
The value-scale condition is kept as an alternative trigger, so the gate can only add subdivisions.

\paragraph{Bracketed emission.}
When an interval with a visibility cue reaches the bracket width $b=2\,\mu$s and its midpoint lies between its endpoint values, the signal is treated as monotone within the bracket and interpolated linearly between the endpoints.
All levels between $L(t_0)$ and $L(t_1)$ are emitted in order at the linearly interpolated times, so each event lies within the bracket and its timing error is at most $b$.
Brackets that are not monotone are refined further by the regular controller.

\section{Sparse Tile Scheduling}
\label{sec:supp_schedule}

\paragraph{Requests and tiles.}
Pixels are grouped into $8\times8$ tiles.
At each refinement level, the controller produces a set of pixels per query time, which is converted into the set of tiles containing at least one requesting pixel.
The renderer projects only Gaussians whose footprints intersect these tiles during the interval and composites only the requested tiles.
To amortize projection, candidate Gaussians are collected for $16\times16$ parent tiles over the time interval and shared by their four child tiles.

\paragraph{Caching.}
Samples are cached by time for the pixels that requested them.
Since every child endpoint coincides with a parent endpoint or midpoint, the samples of an interval remain valid until the interval is resolved, and the cache never re-renders a time it already holds for a pixel.
At most 64 query times are kept, and the times of intervals under refinement are protected from eviction.

\paragraph{Cost accounting.}
The rendering cost counts every queried tile by its area and normalizes by the image area, so a query of all tiles equals one full frame.
$Q$ is the total cost per 0.5\,ms; the base grid alone contributes $Q=1/16$, one full frame every 8\,ms.
For frame-driven baselines, $Q$ equals the number of rendered frames per 0.5\,ms, and for TIDES the number of tangent renders.
A tangent render costs more than an RGB render, so $Q$ favors the baselines, and we report wall time alongside it.

\begin{figure*}[t]
  \centering
  \includegraphics[width=\linewidth]{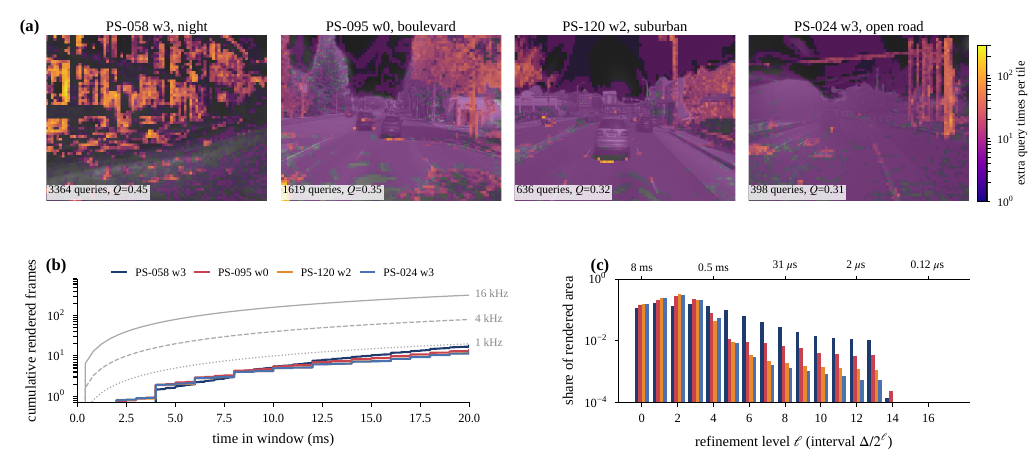}
  \caption{\textbf{Where \method spends its renders} in four 20\,ms ego-and-actor windows at $640\times480$.
  (a) Number of additional query times per $8\times8$ tile beyond the base grid, overlaid on the rendered frame; uncolored tiles are resolved by the base grid alone.
  (b) Cumulative number of full-frame-equivalent renders over the window, compared with frame-driven simulation at 1, 4, and 16\,kHz.
  (c) Share of the rendered area per refinement level.}
  \label{fig:supp_queries}
\end{figure*}

\paragraph{Where the renders go.}
\Cref{fig:supp_queries} visualizes the requests of the scheduler.
Refinement concentrates on tiles with moving edges, thin structures such as poles, foliage, and lit windows, and occlusion boundaries, whereas road surfaces and sky are resolved by the base grid (\cref{fig:supp_queries}a).
Over a 20\,ms window, the four windows render the equivalent of 12.5 to 17.9 full frames, fewer than frame-driven simulation at 1\,kHz renders (20 frames), although their query times reach intervals as short as $\Delta/2^{16}$ (\cref{fig:supp_queries}b).
This is possible because deep levels involve few tiles: levels with intervals of 125\,$\mu$s or less account for 1--5\% of the rendered area in the three daytime windows and for 20\% in the night window PS-058, which contains many small lit structures (\cref{fig:supp_queries}c).
The number of distinct query times varies by an order of magnitude, from 398 to 3364, while the rendered area varies by less than 1.5$\times$, so cost is governed by area rather than by the number of queries.

\section{Implementation Details}
\label{sec:supp_impl}

\paragraph{Hardware and software.}
All experiments run on a server with eight NVIDIA A800 GPUs.
Each simulation uses a single GPU, and independent windows, seeds, and methods are distributed across the eight GPUs.
Wall times in the main comparison are medians over windows on shared GPUs; the accuracy--cost frontier is measured with one process per exclusive GPU, after a warm-up, as the median of three interleaved runs.
Wall times exclude scene loading and include all rendering, solving, and event output.
The renderer is implemented as CUDA kernels in PyTorch and follows the arithmetic of gsplat for the primal pass.

\paragraph{Parameters.}
\Cref{tab:supp_params} lists all parameters of \method.
They are shared by every experiment, except for the time tolerance in the frontier and the M3ED operating points, and the contrast threshold and radiometric response on M3ED.

\begin{table}[t]
\centering
\caption{\textbf{Parameters of \method} used in all experiments unless stated otherwise.}
\label{tab:supp_params}
\footnotesize
\setlength{\tabcolsep}{4pt}
\begin{tabular}{lll}
\toprule
Symbol & Meaning & Value\\
\midrule
$\Delta$ & base interval & 8\,ms\\
$\varepsilon_t$ & time tolerance & 4\,$\mu$s\\
$s$ & safety factor & 2\\
$b$ & bracket width & 2\,$\mu$s\\
$\rho$ & signal-scale ratio & 0.025\\
$h_{\min}$ & minimum interval & $\Delta/2^{16}\approx0.12\,\mu$s\\
-- & tile size & $8\times8$ (parents $16\times16$)\\
$\alpha_{\min}$ & opacity / transmittance cut-off & $10^{-5}$\\
$\epsilon$ & log offset & $10^{-6}$\\
$C$ & contrast threshold (synthetic) & 0.2\\
-- & cached query times & 64\\
\bottomrule
\end{tabular}
\end{table}

\paragraph{Reproducibility.}
All scenes, windows, seeds, and configurations were registered before the corresponding experiments, and the code was frozen during each formal run.
Every result cell is tagged with a signature of its parameters, configuration, assets, and code version, and is never resumed from a cell with a different signature.
Failed cells are archived rather than removed and are reported.
Repeating the M3ED experiments of \method reproduces every event exactly.

\input{tables/supp_tab_synthetic}

\section{Synthetic Benchmark}
\label{sec:supp_synthetic}

\subsection{Protocol}
\label{sec:supp_protocol}

\paragraph{Scenes and windows.}
We use PandaSet scenes 011, 024, 032, 058, 068, 086, 095, and 120, which cover day and night, urban and suburban streets, and parked and moving vehicles.
For each scene, four windows of 20\,ms are selected by quantiles of the image motion, without access to any event result.
Each window is simulated with three motion types.
\emph{Ego} freezes all actors at the window start and keeps the camera motion.
\emph{Actor} freezes the camera and renders only the selected actor, which isolates object motion against an empty background.
\emph{Ego and actor} keeps both motions and all Gaussians, which is the most realistic setting and produces the most occlusions.
Every window is simulated at $346\times260$ and $128\times96$, which gives 192 windows.
The frontier uses the ego-and-actor windows w0 and w1 of scenes 024, 032, 086, and 120.
The ablation uses scenes 011, 058, and 095 with all windows and motion types.

\begin{table}[!htb]
\centering
\caption{\textbf{Macro-averaged F1@5\,$\mu$s} (\%) over windows with at least 500 reference events ($n$ windows) and over all windows.}
\label{tab:supp_synthetic_macro}
\footnotesize
\setlength{\tabcolsep}{3.5pt}
\begin{tabular}{lcccc}
\toprule
& \multicolumn{2}{c}{$346\times260$} & \multicolumn{2}{c}{$128\times96$}\\
\cmidrule(lr){2-3}\cmidrule(lr){4-5}
Method & $\ge$500 ($n$=88) & all & $\ge$500 ($n$=62) & all\\
\midrule
Frame 16\,kHz & 99.836 & 99.069 & 99.897 & 99.885\\
Frame 64\,kHz & 99.884 & 99.170 & 99.940 & 99.935\\
Hermite 31.25\,$\mu$s & 99.860 & 99.061 & 99.924 & 99.911\\
TIDES-adapt.$^\dagger$ & 99.849 & 98.502 & 99.907 & 98.861\\
TIDES-quad.$^\dagger$ & 99.849 & 98.762 & 99.898 & 99.886\\
\method & 99.872 & 98.205 & 99.935 & 99.932\\
\bottomrule
\end{tabular}
\end{table}

\paragraph{Reference.}
The reference renders RGB frames every 3.90625\,$\mu$s (256\,kHz) with an independent projection path, converts them to log luminance, and locates crossings by linear interpolation between consecutive frames.
Its grid is shifted by half a step, so it never coincides with the grid of any frame-driven baseline, and its step is finer than the smallest matching tolerance.

\paragraph{Matching and aggregation.}
Predicted and reference events are grouped by pixel and polarity, and matched by maximum-cardinality matching with minimal total time difference within a tolerance of 5, 10, 31.25, or 62.5\,$\mu$s.
Precision, recall, and F1 are computed from the pooled matches over all windows (event-level F1), which weighs every event equally.
We also report the macro average over windows with at least 500 reference events (\cref{tab:supp_synthetic_macro}); windows with fewer events, such as the three actor-only windows of scene 068 with 9--12 reference events each, are dominated by a handful of events.

\begin{figure*}[t]
  \centering
  \includegraphics[width=\linewidth]{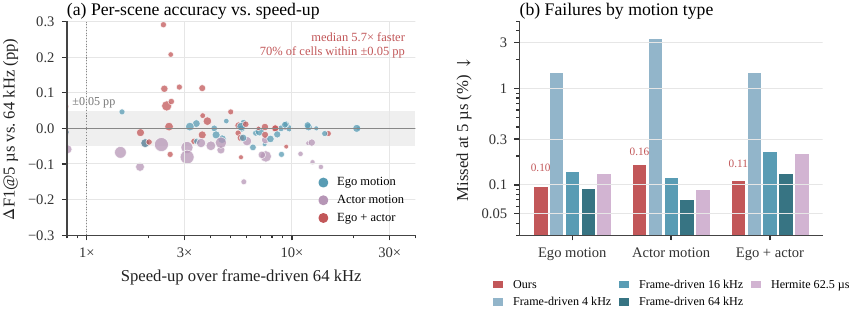}
  \caption{\textbf{Per-window analysis} at $346\times260$. (a) F1 difference to 64\,kHz frames versus speed-up; marker size indicates the event count. (b) Miss rate at 5\,$\mu$s by motion type.}
  \label{fig:supp_perwindow}
\end{figure*}

\subsection{Baselines}
\label{sec:supp_baselines}

\paragraph{Frame-driven simulation.}
The frame-driven baseline renders floating-point RGB frames at a fixed rate with the same renderer, converts them to log luminance, and places one event per crossed level at the linearly interpolated time, as in Vid2E~\cite{gehrig2020vid2e} and ESIM~\cite{rebecq2018esim}.
ESIM, v2e~\cite{hu2021v2e}, and DVS-Voltmeter~\cite{lin2022dvsvoltmeter} are run through their official interfaces on the same floating-point frames; for v2e, we report both the default setting and its HDR option, which takes log-linear input.
Times are kept in double precision to avoid quantization of absolute timestamps.

\paragraph{Fixed-step Hermite interpolation.}
This baseline uses our tangent renderer and the crossing solver of \cref{sec:supp_controller} on a uniform grid of 15.625, 31.25, or 62.5\,$\mu$s, without risk screen, embedded estimate, or bracketing.
It therefore isolates the contribution of error control.

\input{tables/supp_tab_tides}

\paragraph{TIDES.}
TIDES~\cite{thirgood2026tides} is the work closest to ours.
It obtains temporal derivatives by forward-mode differentiation through the deformable scene, predicts possibly multiple crossings per step from a linear model, or a quadratic model when a second render is available, and shortens its step by a factor $\kappa(r)\in(0,1]$ of a risk score $r$ that combines the transmittance rate $|\dot\alpha|$, the contributor count, and the coverage $b$.
An optional tile-level arbiter models finite readout bandwidth, which concerns the sensor model rather than the solver and is therefore not used by any method in our comparison.
As the official code has not been released at the time of writing, we re-implement the linear, quadratic, and adaptive models on our renderer, so that all methods share the same scene and analytic derivatives and differ only in how event times are determined.
The adaptive model evaluates $r=\mathrm{clip}\big(w_b(1-\nu(b))+w_c\,\nu(c)+w_\alpha\,\nu(|\dot\alpha|)\big)$ per pixel with equal weights and $\nu(x)=x/(x+\mathrm{median}(x))$, and scales the base step by $\kappa(R)=1-(1-\kappa_{\min})R$, where $R$ is the 0.99 quantile of $r$ over the image.
Compared with \method, TIDES neither estimates the timing error of an emitted event nor accepts a time tolerance, its steps shrink according to risk rather than to an error estimate, and it renders the full image at every query.
Parameters that the paper leaves unspecified, namely the weights, $\kappa_{\min}=0.1$, the risk quantile, the emission threshold $b_{\min}=10^{-3}$ on the accumulated opacity, and at most 64 events per pixel and step, were fixed before any experiment.
The published method suppresses events at pixels whose accumulated opacity is below $b_{\min}$, which removes faint actor edges in front of empty background.
\Cref{tab:supp_tides} shows that evaluating the gate at either end of a step does not change the result, whereas removing it raises the F1 from about 87\% to the values reported in the main paper; a variant that additionally emits crossings missed by the linear prediction but revealed by the next render gives a small further gain.
We report the variants without the gate in the main paper, which strengthens the baseline.

\input{tables/supp_tab_ablation}

\subsection{Main Results}
\label{sec:supp_results}

\Cref{tab:supp_synthetic} lists all methods at both resolutions and four matching tolerances.
At $128\times96$, \method reaches 99.905\% F1 at 5\,$\mu$s in 1.66\,s, compared with 99.934\% for 64\,kHz frames in 23.8\,s, which mirrors the results at $346\times260$.
\Cref{tab:supp_synthetic_macro} reports macro averages: over windows with at least 500 reference events, \method is within $1.2\times10^{-4}$ of the best configuration at both resolutions.
Including all windows, the three actor-only windows of scene 068, each with fewer than 15 reference events, lower the macro average of \method at $346\times260$, which reflects these near-empty windows rather than the bulk of events.

\paragraph{Per-window behavior.}
\Cref{fig:supp_perwindow}a plots, for each window with at least 500 reference events, the F1 difference between \method and 64\,kHz frame-driven simulation against the speed-up.
In 70\% of the windows, the difference lies within $\pm0.05$ percentage points, at a median speed-up of 5.7$\times$, which shows that the parity in \cref{tab:synthetic} holds per window rather than only on average.
\Cref{fig:supp_perwindow}b separates the miss rate by motion type.
With both ego and actor motion, which produces the most occlusions and is the typical driving case, \method misses the fewest events of all methods; on actor-only windows, a few distant actors yield very sparse windows that dominate the remaining difference.

\begin{figure*}[t]
  \centering
  \includegraphics[width=\linewidth]{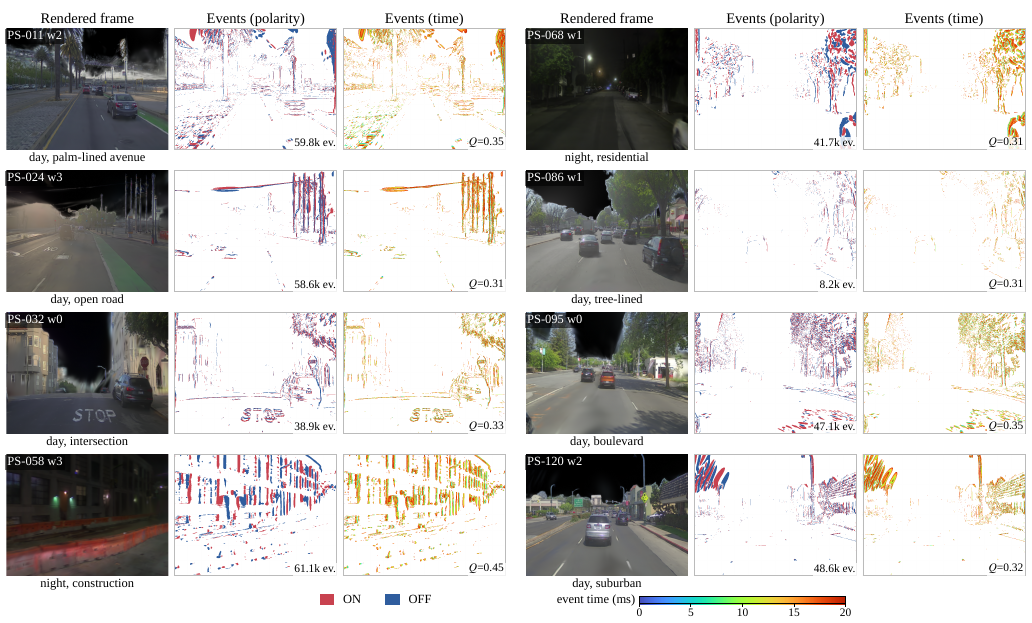}
  \caption{\textbf{Event streams of \method on all eight PandaSet scenes} (ego and actor, $640\times480$, 20\,ms, windows not shown in the main paper).
  For each scene: the rendered frame at the window center, the events colored by polarity (latest event per pixel), and the same events colored by time.
  Badges give the number of events and the rendering cost $Q$ (full frames per 0.5\,ms).}
  \label{fig:supp_gallery}
\end{figure*}

\begin{figure*}[t]
  \centering
  \includegraphics[width=\linewidth]{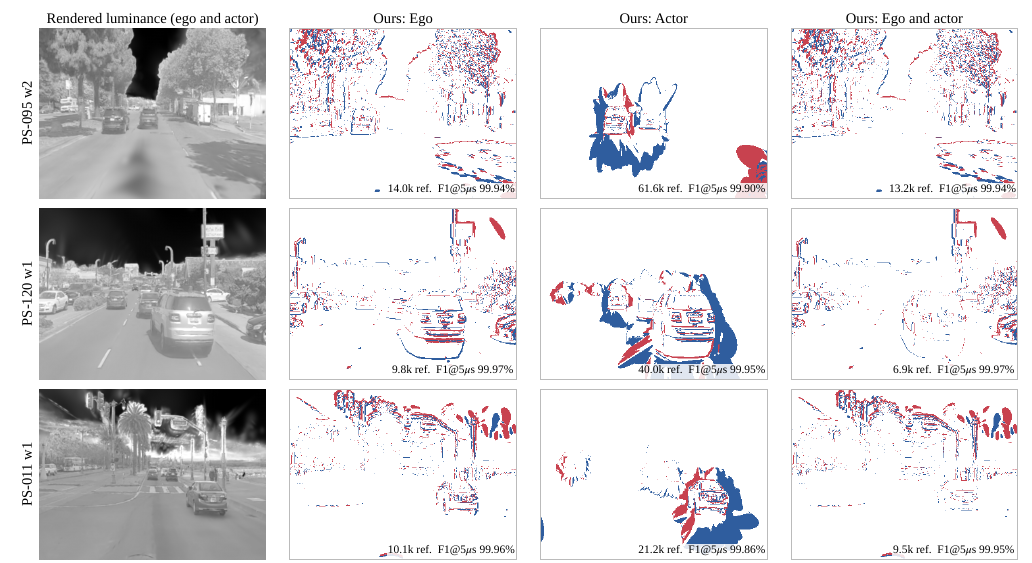}
  \caption{\textbf{The three motion types} of the synthetic benchmark on three windows at $346\times260$.
  \emph{Ego} freezes the actors, \emph{Actor} freezes the camera and keeps only the actor Gaussians in front of an empty background, and \emph{Ego and actor} keeps both motions and all Gaussians.
  Badges give the number of reference events and the F1@5\,$\mu$s of \method.}
  \label{fig:supp_motion}
\end{figure*}

\subsection{Qualitative Results}
\label{sec:supp_qualitative}

\Cref{fig:supp_gallery} shows the events of \method on one window of every synthetic scene, rendered at $640\times480$ to make fine structures visible.
The scenes cover day and night, open roads, intersections, and tree-lined streets; in all of them, the events outline vehicles, lane markings, poles, and facades, and the time-colored panels show that the events of a moving edge form a continuous sweep rather than discrete copies of the edge, as a frame-driven simulator with a low rate would produce.
The rendering cost stays between $Q=0.31$ and $0.45$, \ie 0.6 to 0.9 full frames per millisecond, although the event counts differ by a factor of 7.5.

\Cref{fig:supp_motion} separates the motion types.
Ego motion produces events on the whole static background, whereas actor motion alone produces events only on the actor; since the actor-only setting removes the background, the semi-transparent outer Gaussians of an actor sweep over an empty image and generate large regions of events with small contrast steps, which are among the hardest cases for any sampler.
Ego-and-actor windows combine both and add occlusions between actors and background.
\method reaches an F1@5\,$\mu$s between 99.86\% and 99.97\% on all nine windows shown.

\subsection{Distribution of Timing Errors}
\label{sec:supp_timing}

\begin{figure*}[t]
  \centering
  \includegraphics[width=\linewidth]{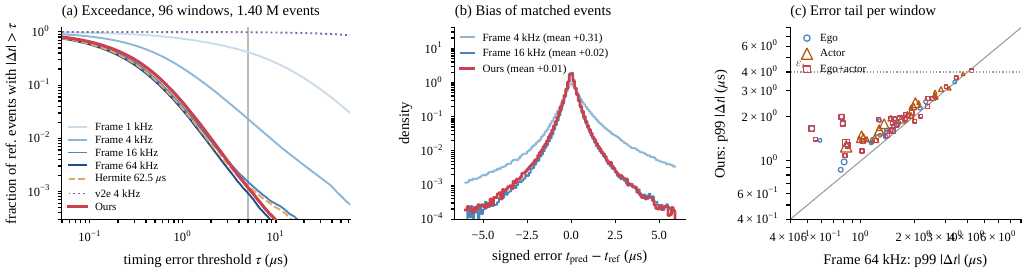}
  \caption{\textbf{Timing errors over all 96 windows at $346\times260$.}
  Events are matched to the reference within 62.5\,$\mu$s as in the main benchmark; unmatched reference events count as errors larger than every threshold.
  (a) Fraction of reference events whose timing error exceeds $\tau$; the value at 5\,$\mu$s is the miss rate behind the recall at 5\,$\mu$s.
  (b) Distribution of the signed error of matched events.
  (c) 99th percentile of the absolute error per window ($\ge$500 reference events), \method versus 64\,kHz frames; the dotted line marks $\varepsilon_t=4\,\mu$s.}
  \label{fig:supp_timing}
\end{figure*}

F1 at a fixed tolerance summarizes the timing error by a single threshold.
\Cref{fig:supp_timing}a shows the complete distribution over the 1.40\,M reference events of the main benchmark at $346\times260$.
The curve of \method lies on those of 16\,kHz and 64\,kHz frames and of Hermite interpolation with fixed steps, \ie of the methods that sample the signal densely everywhere: the median absolute error is 0.17\,$\mu$s, the 99th percentile 2.0\,$\mu$s, and 0.12\% of the reference events are matched with an error above 5\,$\mu$s or not at all, compared with 0.15\% for 16\,kHz and 0.09\% for 64\,kHz frames.
At lower rates, the tail grows quickly: with 4\,kHz frames, 2.3\% of the events deviate by more than 5\,$\mu$s, and with 1\,kHz frames 42\%.
ESIM produces the same curves as the frame-driven baseline, because both interpolate linearly between the same floating-point frames, and v2e, in the official configuration used here, assigns every event the time of one of its input frames, so that fewer than 14\% are matched within 62.5\,$\mu$s.
\Cref{fig:supp_timing}b reveals a systematic bias of frame-driven simulation: matched events are late by $+0.31\,\mu$s on average at 4\,kHz and $+3.0\,\mu$s at 1\,kHz, because linear interpolation between frames misplaces crossings of curved signals in a consistent direction, whereas the mean error of \method is $+0.01\,\mu$s.
\Cref{fig:supp_timing}c shows that the error tail of \method follows that of 64\,kHz frames window by window, and that its 99th percentile stays below the requested tolerance $\varepsilon_t=4\,\mu$s in all but one of the 88 windows, which reaches 4.1\,$\mu$s.

\subsection{Behavior at Single Pixels}
\label{sec:supp_pixels}

\begin{figure*}[t]
  \centering
  \includegraphics[width=\linewidth]{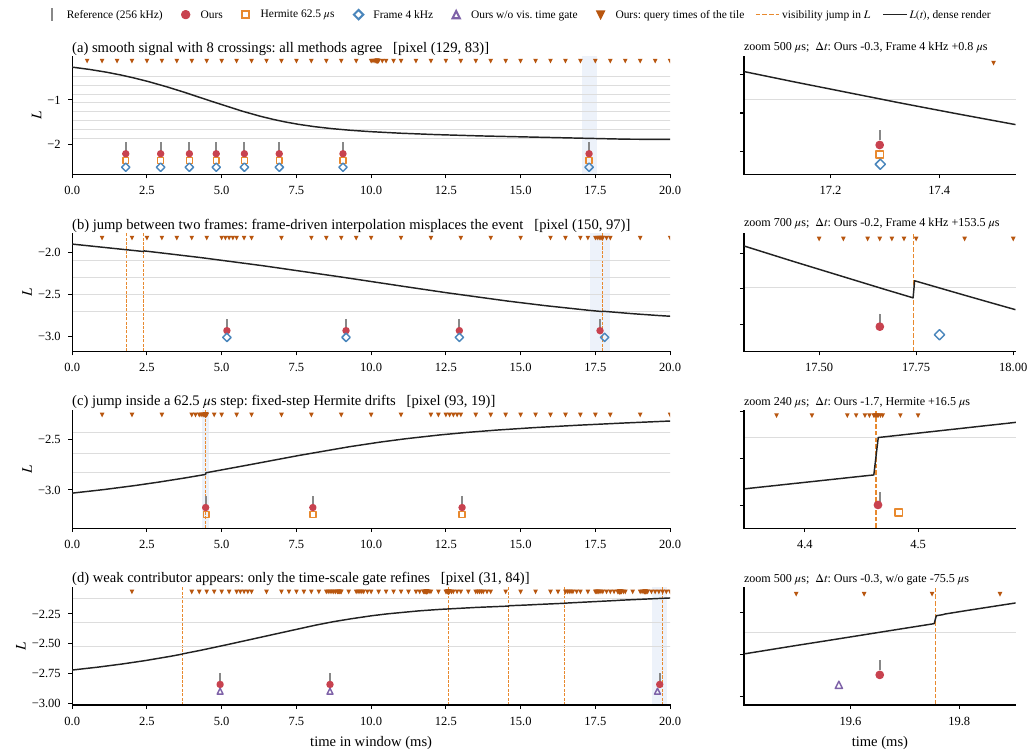}
  \caption{\textbf{Four pixels of window PS-058 w3} (ego and actor, $346\times260$).
  The black curve is the log brightness rendered every 3.9\,$\mu$s; horizontal lines mark the crossed levels; markers below the curve are the events of each method, markers above the curve the query times of \method for the tile that contains the pixel; dashed lines mark jumps of $L$ caused by visibility changes.
  The right column zooms into the shaded interval, and its title gives the timing errors of \method and of the method that fails at this pixel.}
  \label{fig:supp_pixels}
\end{figure*}

\Cref{fig:supp_pixels} follows four pixels through a night window and illustrates, on real renders, the mechanisms discussed in \cref{sec:supp_controller,sec:supp_visibility}.
(a)~Where $L$ is smooth, all methods agree to within a microsecond.
(b)~A visibility jump occurs 90\,$\mu$s after a crossing but within the same frame interval of a 4\,kHz frame-driven simulation; linear interpolation between the two frames then places the event 154\,$\mu$s too late, although the crossing itself happened on a smooth part of the signal.
(c)~An abrupt rise of $L$ by 0.016, caused by a visibility change, falls inside one 62.5\,$\mu$s step; the cubic of fixed-step Hermite interpolation smooths it and shifts the event by 16.5\,$\mu$s, whereas the embedded estimate of \method detects the disagreement and refines down to the rise.
(d)~A weak contributor changes $L$ by a step of $9\times10^{-4}$, 100\,$\mu$s after a crossing.
The step is about a fifth of $\rho C=5\times10^{-3}$, so the value-scale condition does not trigger, and without the time-scale gate the solver accepts a cubic that is displaced by 76\,$\mu$s; with the gate, the timing defect $\delta/\sigma$ forces refinement and the error falls to 0.3\,$\mu$s.

\input{tables/supp_tab_frontier}

\subsection{Ablation}
\label{sec:supp_ablation}

\Cref{tab:supp_ablation} completes the ablation of the main paper with both resolutions and the optional safeguards.
Quarter probes, the double-precision fallback, and the precision guard leave the F1 unchanged within $3\times10^{-5}$ on these scenes; we keep them as safety mechanisms rather than as contributions.

\begin{figure*}[t]
  \centering
  \includegraphics[width=\linewidth]{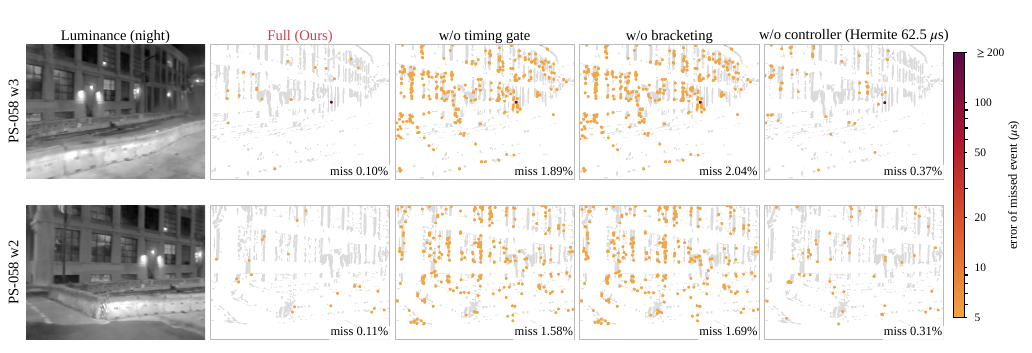}
  \caption{\textbf{Events lost by ablated components} on two night windows of the ablation scenes (ego and actor, $346\times260$).
  Gray: reference events; colored: reference events missed at 5\,$\mu$s, colored by the timing error of the nearest event of the same pixel and polarity.
  The last column (w/o controller) replaces error control by Hermite interpolation with a fixed step of 62.5\,$\mu$s from the main benchmark.}
  \label{fig:supp_ablation_maps}
\end{figure*}

\Cref{fig:supp_ablation_maps} shows where the ablated components matter.
Without the timing gate, and slightly more without bracketing, misses at 5\,$\mu$s rise from about 0.1\% to 1.6--2.0\% in the two night windows of scene 058, which are among the ablation windows with the largest degradation.
The missed events lie on lit facades and window frames, where many faint Gaussians overlap, and are displaced by 5--20\,$\mu$s, which matches the mechanism of \cref{sec:supp_visibility} and \cref{fig:supp_pixels}d.
Hermite interpolation with fixed steps of 62.5\,$\mu$s misses 0.31--0.37\% of the events, mostly in the same regions; the embedded estimate of \method removes most of these misses.
Sparse queries and the tile size change the cost but leave the F1 of these windows unchanged, so their maps are not shown.

\input{tables/supp_tab_calibration}

\subsection{Accuracy--Cost Frontier}

\Cref{tab:supp_frontier} gives all frontier configurations.
\method with $\varepsilon_t=8\,\mu$s matches 32\,kHz frames at 4.8$\times$ less time, and $\varepsilon_t=4\,\mu$s exceeds 64\,kHz frames and 15.625\,$\mu$s Hermite interpolation at 8.5$\times$ and 12.9$\times$ less time.
At the lowest accuracy, $\varepsilon_t=16\,\mu$s and 8\,kHz frames are comparable, as the fixed overhead per query then dominates the cost of \method.

\input{tables/supp_tab_m3ed}

\input{tables/supp_tab_m3ed_protocols}

\subsection{Calibration of the Time Tolerance and Reference Accuracy}
\label{sec:supp_calibration}

Since the embedded estimate is not a strict bound, we measure how well the tolerance $\varepsilon_t$ controls the realized error.
We re-simulate four ego-and-actor frontier windows (w0 and w1 of scene 024, and w0 of scenes 032 and 120) with $\varepsilon_t\in\{1,2,4,8,16\}\,\mu$s and compare all events with a reference rendered every 0.98\,$\mu$s, four times finer than the reference of the main benchmark.
\Cref{tab:supp_calibration} shows three findings.
First, the standard reference with a 3.9\,$\mu$s step agrees with the finer one to a median of 0.13\,$\mu$s, and 99\% of its events lie within 1\,$\mu$s; none deviates by more than 5\,$\mu$s, so the matching tolerances of the main paper are not limited by the reference.
Second, the tolerance is well calibrated for $\varepsilon_t\ge2\,\mu$s: at the default $\varepsilon_t=4\,\mu$s, only 0.27\% of the events deviate by more than $\varepsilon_t$ and 0.06\% by more than $2\varepsilon_t$, and these fractions decrease monotonically as $\varepsilon_t$ grows.
At $\varepsilon_t=1\,\mu$s, the requested accuracy approaches the floor set by single-precision rendering and the reference itself, and more events exceed the tolerance.
Third, against the finer reference, \method with $\varepsilon_t=4\,\mu$s misses fewer events at 5\,$\mu$s than 64\,kHz frame-driven simulation (0.16\% versus 0.17\%), so the parity reported in the main paper is not an artifact of the reference resolution.
The remaining deviations concentrate in scene 032, where both references also disagree more often, which indicates fine structures close to the resolution of all samplers rather than a failure specific to the controller.

\section{Real Recordings}
\label{sec:supp_real}

\subsection{Protocol}
\label{sec:supp_m3ed_protocol}

\paragraph{Windows and reconstruction.}
We use the \emph{car\_urban\_day\_horse} and \emph{car\_urban\_day\_rittenhouse} sequences at the full $1280\times720$ resolution of the event camera.
Each sequence has 12 registered windows of 50\,ms, spread over the sequence and away from its ends, and each window is preceded by a 10\,ms warm-up that is not evaluated.
For each window, the scene is refined from a global reconstruction using RGB frames and events within a 1.2\,s context, excluding all data within $\pm100$\,ms of the window.
Static and dynamic candidates are refined, and one is selected by a fixed photometric rule that never accesses test events; a window without a valid candidate would count as a failure, which did not occur.
Camera trajectories combine ground-truth poses with gyroscope measurements and are refined with a spline with knots every 80\,ms.

\begin{figure*}[t]
  \centering
  \includegraphics[width=\linewidth]{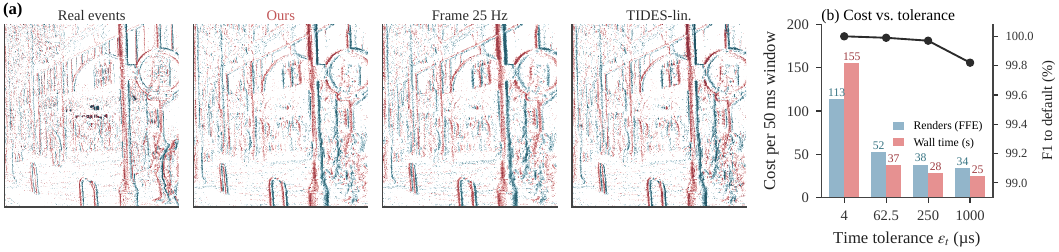}
  \caption{\textbf{M3ED.} (a) Time surfaces of the last 10\,ms of a window; all simulators render the same scene, so their spatial structure is similar, and the real recording additionally contains sensor noise and a pedestrian absent from the reconstruction. (b) Renders and wall time per window, and agreement with the default setting, for different time tolerances.}
  \label{fig:supp_m3ed}
\end{figure*}

\paragraph{Seeds and calibration.}
Real sensors start from unknown reference levels.
Each phase seed draws the initial reference of every pixel uniformly within one contrast step, identically for all simulators, and results are summarized by the median over three seeds and then averaged over windows.
The radiometric response and the contrast threshold are calibrated on non-test windows by matching the total event count and the event-count map.

\paragraph{Protocols.}
The main protocol follows TIDES and renders frame-driven baselines at ten times the camera rate (250\,Hz), and TIDES with the same step.
We additionally report the native camera rate (25\,Hz), a budget protocol in which frame-driven baselines render the same image area as \method, and a dense protocol at 4\,kHz, which approximates the continuous-time limit of the scene.

\paragraph{Metrics.}
Voxel-grid and count-map similarities are cosine similarities between voxel grids~\cite{zhu2019unsupervised} and per-pixel event counts, and the time-surface similarity uses exponentially decaying time surfaces~\cite{lagorce2017hots}.
The count error is the relative difference in the total event count.
The Chamfer distance is computed in $(x,y,t)$ with the time axis scaled by the median image speed of the window, which is obtained from the rendered depth and the camera motion.
The log-ISI distance is the 1-Wasserstein distance between the distributions of log inter-event intervals at the same pixel, and the Fano error is the log ratio of the Fano factors of per-pixel counts.
Scene fidelity matches the events of a method with the 4\,kHz dense events of the same scene at tolerances of 0.25 and 1\,ms.

\subsection{Results}

\Cref{tab:supp_m3ed} reports all real-data metrics under the main protocol, and \cref{tab:supp_m3ed_protocols} the remaining protocols.
With the same rendered area (budget) or dense 4\,kHz frames, frame-driven simulation converges to the same continuous signal as \method, and all metrics agree to the third decimal place.
At the native camera rate, by contrast, events are misplaced in time, and the scene fidelity drops to about 0.1.
\input{tables/supp_tab_fidelity}

\Cref{tab:supp_fidelity} details the scene-fidelity analysis.
All simulators that are faithful to the scene produce about 68\% of the real event count, as reconstructions are smoother than the real world and contain no sensor noise.
TIDES-linear exceeds the scene's own event count by 5.6\% and produces opposite-polarity events within 2\,ms, a signature of overshooting extrapolation that is absent in the scene and reduces its fidelity to 0.59.

\paragraph{Operating points.}
As shown in \cref{fig:supp_m3ed}b and \cref{tab:supp_m3ed_protocols}, relaxing $\varepsilon_t$ from 4\,$\mu$s to 62.5\,$\mu$s, 250\,$\mu$s, and 1\,ms reduces the renders per window from 113 to 52.5, 38, and 34.5 and the wall time by up to 6.3$\times$, while 99.8--99.99\% of the events coincide with the default setting and all real-data metrics remain unchanged.

\begin{figure*}[t]
  \centering
  \includegraphics[width=\linewidth]{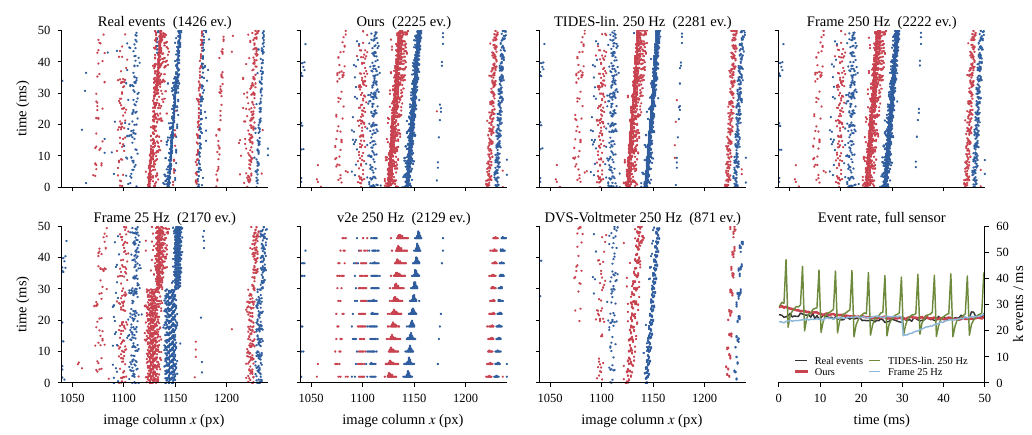}
  \caption{\textbf{Space--time structure on M3ED} (\emph{car\_urban\_day\_horse}, window e3, phase seed 0, the window of \cref{fig:supp_m3ed}a).
  Each panel shows the events of rows 159--161 and columns 1040--1240 over the 50\,ms window, with time on the vertical axis; a moving edge appears as a slanted band.
  Baselines run at ten times the camera rate unless stated otherwise.
  Bottom right: event rate of the full sensor in bins of 0.5\,ms.}
  \label{fig:supp_m3ed_xt}
\end{figure*}

\paragraph{Space--time structure.}
\Cref{fig:supp_m3ed_xt} complements the aggregate metrics with a direct view of the event streams.
In the real recording, a facade edge moves by about 15 pixels within the window and produces two parallel slanted bands of opposite polarity.
\method, TIDES, and 250\,Hz frames reproduce this band, its slope, and the separation of the polarities.
At the native 25\,Hz, the window spans only one frame boundary, at about 30\,ms; the interpolated edge moves at a constant speed within each frame interval and changes its speed abruptly at the boundary, so the band breaks into two straight segments.
v2e assigns all events to the times of its input frames, which shows as horizontal stripes every 4\,ms, and DVS-Voltmeter, whose stochastic model emits fewer events on this scene, keeps only the strongest edges.
The event rate of the full sensor (bottom right) is nearly constant for the real recording and for \method, with a coefficient of variation over 0.5\,ms bins of 0.036 and 0.049.
TIDES-linear shows one peak in every 4\,ms step, with a coefficient of variation of 0.24 and peaks 1.8 times the median rate, a temporal counterpart of the overshooting extrapolation discussed above.

\section{Sim-to-Real Optical Flow}
\label{sec:supp_flow}

\paragraph{Data.}
Training data come from the \emph{car\_urban\_day\_penno\_small\_loop} and \emph{car\_urban\_day\_horse} sequences, and test data from \emph{car\_urban\_day\_rittenhouse}, all within a $320\times256$ region of the event sensor without resampling.
Segments of 0.6\,s are taken on a regular grid; every fourth training segment is held out for validation.
Test segments are at least 1\,s away from all registered event-quality windows and from five development segments, which were inspected during development and are reported separately as dev-test.
Training labels are computed from the rendered depth and the ground-truth poses and describe the static-world flow over 50\,ms.
Test labels are derived from the global LiDAR map, projected with the ground-truth poses and filtered for occlusion, and are therefore sparse.

\paragraph{Network and training.}
The input stacks two consecutive 50\,ms windows as a 10-bin voxel grid.
A U-Net~\cite{ronneberger2015unet} with base width 32 regresses the flow and is trained with an L1 end-point loss for 8000 iterations with batch size 16 and learning rate $3\times10^{-4}$.
The checkpoint with the lowest EPE on the simulated validation set is evaluated.
Each training input is used with five seeds, and two controls are included: an input without events and an input with spatially shuffled events.
Before any comparison, we required that training on real events reduces the dev-test EPE by more than 20\% relative to the input without events; it reduced the EPE by 30\%.
Training the same configuration twice reproduces identical weights and metrics.

\input{tables/supp_tab_flow}

\paragraph{Results.}
\Cref{tab:supp_flow} lists all training inputs.
All continuous-time inputs, including 250\,Hz and 4\,kHz frame-driven simulation, TIDES, and \method, reach the level of training on real events, whereas inputs at the native camera rate and DVS-Voltmeter, whose stochastic events deviate from the scene, remain clearly worse.
The dev-test segments show the same ordering with smaller differences, as their image motion is slower.

\begin{figure*}[t]
  \centering
  \includegraphics[width=\linewidth]{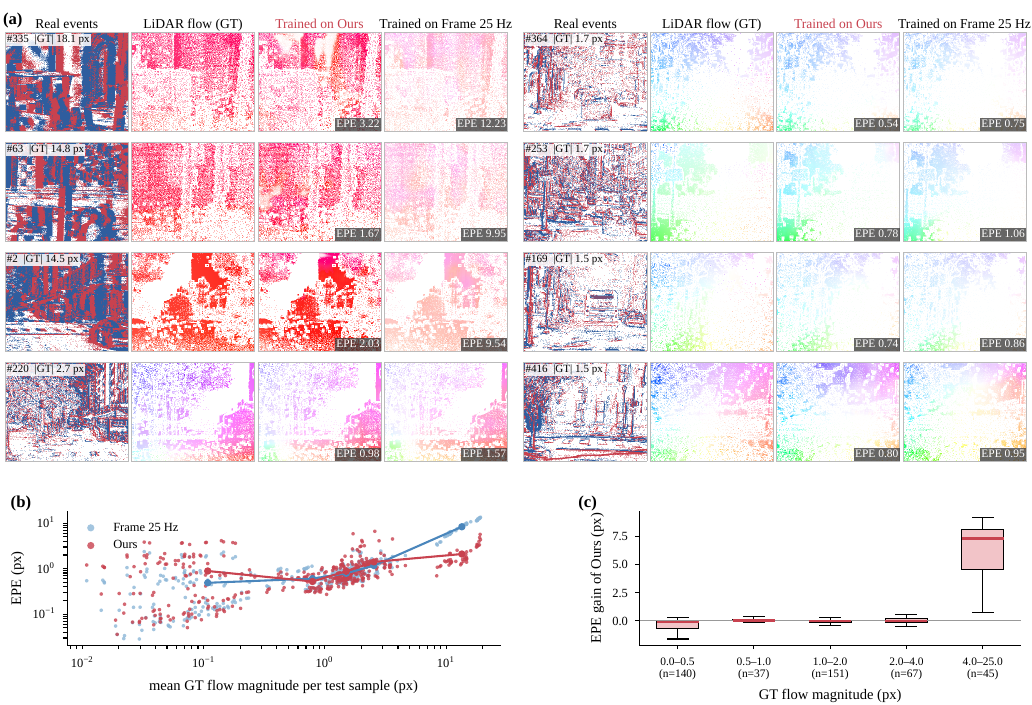}
  \caption{\textbf{Additional optical-flow results on real M3ED test events} (the checkpoints of \cref{fig:flow}, one seed).
  (a) Eight test samples not shown in the main paper, ordered by the mean magnitude of the ground-truth flow; flow is color-coded relative to the 99th percentile of the ground truth and shown only on the LiDAR support, and badges give the per-sample EPE.
  (b) Per-sample EPE of all 440 test samples versus the ground-truth flow magnitude, with means per magnitude bin.
  (c) EPE gain of the network trained on \method over the one trained on 25\,Hz frames, per magnitude bin.}
  \label{fig:supp_flow}
\end{figure*}

\paragraph{Breakdown by motion magnitude.}
\Cref{fig:supp_flow} breaks down the result of the networks trained on \method and on 25\,Hz frames.
For the 45 test samples with a mean flow above 4 pixels per 50\,ms, the network trained on \method is better on every sample, with a mean EPE of 2.09 compared with 8.29; at these speeds, 25\,Hz frames sample an edge only a few times along its path, and the simulated events no longer resemble those of the real sensor.
For flow between 0.5 and 4 pixels, the two networks are within 0.25 of each other (mean EPE of \method versus 25\,Hz frames: 0.52 versus 0.60, 0.88 versus 0.79, and 1.40 versus 1.17 in the three bins).
For nearly static samples below 0.5 pixels, the network trained on 25\,Hz frames is better (0.49 versus 0.89 for \method).
The overall advantage of \method therefore stems from fast motion, which is exactly where event cameras are used and where the temporal resolution of the simulator matters.
The qualitative examples in \cref{fig:supp_flow}a show the same pattern: for fast samples, the network trained on 25\,Hz frames underestimates the flow and predicts pale colors, whereas both networks recover the flow field of slow samples.

\section{Long Sequences}
\label{sec:supp_long}

\input{tables/supp_tab_long}

The long-sequence experiment uses all Gaussians of 8 PandaSet and 8 nuScenes scenes in the ego-and-actor setting at $346\times260$, without any subset selection.
Each run covers 0.5 or 5\,s and is processed in chunks of 0.1\,s, between which the reference levels and the last samples are carried over, so the output equals a single continuous run.
Since no dense reference is available at this length, the experiment evaluates stability, cost, and memory.

\Cref{tab:supp_long} lists the 5\,s runs per scene.
The mean cost ranges from $Q=0.26$ for the nearly static nuScenes scene 189 to 0.56, and the 95th percentile over base intervals stays below 1.3, so 95\% of the base intervals render less than about 2.6 full frames per millisecond.
Memory is dominated by the number of Gaussians and stays below 10.3\,GB.

\begin{figure*}[t]
  \centering
  \includegraphics[width=\linewidth]{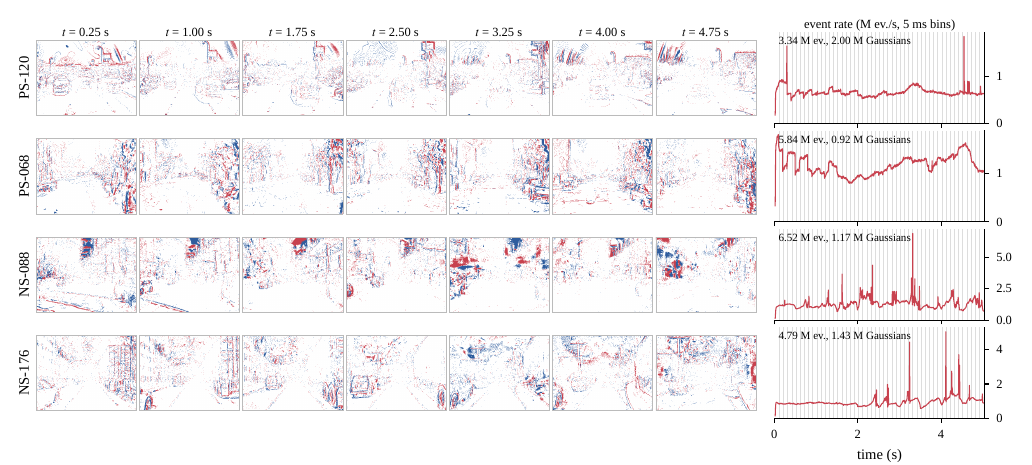}
  \caption{\textbf{Complete 5\,s sequences} of two PandaSet (PS) and two nuScenes (NS) scenes with all Gaussians (ego and actor, $346\times260$).
  Left: events within 10\,ms at seven times. Right: event rate over the whole sequence in 5\,ms bins; thin vertical lines mark the boundaries of the 0.1\,s chunks.}
  \label{fig:supp_long}
\end{figure*}

\Cref{fig:supp_long} shows snapshots and event rates of four complete sequences with up to 2.0\,M Gaussians and up to 6.5\,M events.
The event rate follows the scene content, with slow variations over seconds and short bursts when large structures sweep across the image, and it shows no discontinuity at the 49 chunk boundaries, as expected from carrying the reference levels and samples over between chunks.

\section{Discussion and Limitations}
\label{sec:supp_discussion}

\paragraph{Empirical error control.}
The embedded estimate and the visibility gate are empirical rather than strict bounds.
A pulse that starts and ends between samples without affecting values, derivatives, or visibility cues can be missed, as with any sampling-based solver.
Quarter probes for flat intervals and the visibility cue reduce this risk, and the remaining discrepancies in our diagnostics were dominated by the reference itself, which missed narrow crossings that \method resolved.

\paragraph{Cost on real data.}
On M3ED, the default tolerance of 4\,$\mu$s renders more image area than frame-driven simulation at 250\,Hz, because real recordings are evaluated at millisecond scales.
Relaxing the tolerance closes most of this gap without changing the output, but a cost advantage on real data at coarse tolerances is not claimed.

\paragraph{Sensor realism.}
\method models an ideal sensor, and all simulators share the event-count gap caused by smooth reconstructions.
Noise, refractory periods, threshold variation, and bandwidth limits can be applied to the ideal stream, and their calibration on non-test windows is a natural extension.

%% file: tables/supp_tab_synthetic.tex
\begin{table*}[t]
\centering
\caption{\textbf{Event timing accuracy on synthetic driving scenes.} F1 (\%) of event matching at time tolerances of 5, 10, and 31.25\,$\mu$s against an independent dense reference, median wall time per 20\,ms window, and rendering cost $Q$ (full-frame renders per 0.5\,ms). Results are pooled over 96 windows per resolution (8 PandaSet scenes $\times$ 4 windows $\times$ 3 motion types). $^\dagger$TIDES with its emission gate removed, see \cref{sec:supp_baselines}. \colorbox{rkA}{\textbf{First}}, \colorbox{rkB}{second}, and \colorbox{rkC}{third} per column.}
\label{tab:supp_synthetic}
\footnotesize
\setlength{\tabcolsep}{3.6pt}
\begin{tabular}{llc ccccc cc}
\toprule
& & & \multicolumn{5}{c}{$346\times260$} & \multicolumn{2}{c}{$128\times96$}\\
\cmidrule(lr){4-8}\cmidrule(lr){9-10}
Type & Method & Rate/step & F1@5\,$\mu$s$\uparrow$ & F1@10\,$\mu$s$\uparrow$ & F1@31\,$\mu$s$\uparrow$ & Time (s)$\downarrow$ & $Q\downarrow$ & F1@5\,$\mu$s$\uparrow$ & Time (s)$\downarrow$\\
\midrule
Frame-driven & RGB~\cite{gehrig2020vid2e} & 4 kHz & 97.707 & 99.206 & 99.847 & \fst{1.19} & \snd{2.00} & 98.548 & \fst{1.55} \\
 & RGB~\cite{gehrig2020vid2e} & 16 kHz & 99.850 & 99.942 & 99.988 & \trd{4.53} & \trd{8.00} & 99.891 & 6.02 \\
 & RGB~\cite{gehrig2020vid2e} & 64 kHz & \fst{99.908} & \fst{99.979} & \fst{99.996} & 18.07 & 32.00 & \fst{99.934} & 23.80 \\
 & ESIM~\cite{rebecq2018esim} & 4 kHz & 97.708 & 99.207 & 99.848 & 9.63 & \snd{2.00} & 98.550 & 5.70 \\
 & v2e~\cite{hu2021v2e} & 4 kHz & 1.221 & 2.446 & 7.652 & 11.57 & \snd{2.00} & 0.974 & 8.07 \\
 & v2e-HDR~\cite{hu2021v2e} & 4 kHz & 2.541 & 5.102 & 15.738 & 7.27 & \snd{2.00} & 2.394 & \trd{5.46} \\
\midrule
Fixed step & Hermite & 62.5\,$\mu$s & 99.871 & 99.950 & 99.992 & 6.53 & \trd{8.00} & 99.893 & 10.27 \\
 & Hermite & 31.25\,$\mu$s & \snd{99.889} & \trd{99.967} & \snd{99.995} & 13.54 & 16.00 & \snd{99.920} & 19.86 \\
\midrule
Derivative & TIDES-lin. (gated)~\cite{thirgood2026tides} & 16 kHz & 86.992 & 87.095 & 87.158 & 6.85 & 8.03 & 82.139 & 9.32 \\
 & TIDES-lin.$^\dagger$~\cite{thirgood2026tides} & 16 kHz & 99.713 & 99.879 & 99.961 & 6.96 & 8.03 & 99.802 & 9.92 \\
 & TIDES-quad.$^\dagger$~\cite{thirgood2026tides} & 16 kHz & 99.869 & 99.949 & 99.989 & 7.37 & 8.03 & 99.892 & 9.70 \\
 & TIDES-adapt.$^\dagger$~\cite{thirgood2026tides} & 16 kHz & 99.875 & 99.953 & 99.990 & 22.66 & 23.24 & 99.903 & 29.18 \\
\midrule
 & \textbf{\method (Ours)} & $\varepsilon_t$=4\,$\mu$s & \trd{99.881} & \snd{99.973} & \trd{99.994} & \snd{3.55} & \fst{0.32} & \trd{99.905} & \snd{1.66} \\
\bottomrule
\end{tabular}
\end{table*}

%% file: tables/supp_tab_tides.tex
\begin{table}[!htb]
\centering
\caption{\textbf{TIDES variants} on the main benchmark (96 windows per resolution, event-level F1@5\,$\mu$s in \%). All variants run at 16\,kHz on our renderer.}
\label{tab:supp_tides}
\footnotesize
\setlength{\tabcolsep}{4pt}
\begin{tabular}{llcc}
\toprule
Model & Emission gate & $346\times260$ & $128\times96$\\
\midrule
Linear & gated (start of step) & 86.992 & 82.139 \\
Linear & gated (either end) & 87.010 & 82.157 \\
Linear & no gate & 99.713 & 99.802 \\
Linear & no gate + catch-up & 99.727 & 99.816 \\
\midrule
Quadratic & gated & 86.761 & 81.754 \\
Quadratic & no gate & 99.869 & 99.892 \\
\midrule
Adaptive & gated & 87.077 & 82.191 \\
Adaptive & no gate & 99.875 & 99.903 \\
\midrule
\method & -- & 99.881 & 99.905 \\
\bottomrule
\end{tabular}
\end{table}

%% file: tables/supp_tab_ablation.tex
\begin{table*}[t]
\centering
\caption{\textbf{Complete ablation} on 36 held-out windows per resolution (scenes 011, 058, 095). F1 in \%, median wall time in s. The last three rows are optional safeguards that do not change the results on our benchmarks.}
\label{tab:supp_ablation}
\footnotesize
\setlength{\tabcolsep}{4pt}
\begin{tabular}{l cccc cccc}
\toprule
& \multicolumn{4}{c}{$346\times260$} & \multicolumn{4}{c}{$128\times96$}\\
\cmidrule(lr){2-5}\cmidrule(lr){6-9}
Variant & F1@5 & F1@10 & Time & $Q$ & F1@5 & F1@10 & Time & $Q$\\
\midrule
\method (full) & 99.880 & 99.969 & 3.80 & 0.33 & 99.875 & 99.963 & 2.11 & 0.32 \\
\midrule
w/o visibility bracketing & 99.492 & 99.815 & 2.48 & 0.31 & 99.456 & 99.780 & 1.28 & 0.30 \\
w/o timing gate (\cref{eq:gate}) & 99.567 & 99.866 & 2.34 & 0.31 & 99.494 & 99.801 & 1.37 & 0.30 \\
w/o sparse tile scheduling & 99.880 & 99.969 & 6.33 & 7.03 & 99.875 & 99.963 & 2.27 & 2.23 \\
$16\times16$ tiles & 99.880 & 99.969 & 3.72 & 0.44 & 99.875 & 99.963 & 1.83 & 0.40 \\
\midrule
w/o quarter probes & 99.880 & 99.968 & 3.39 & 0.26 & 99.875 & 99.963 & 1.49 & 0.23 \\
w/o double-precision fallback & 99.883 & 99.972 & 3.48 & 0.33 & 99.876 & 99.965 & 1.60 & 0.32 \\
w/o precision guard & 99.883 & 99.972 & 3.49 & 0.33 & 99.876 & 99.965 & 1.58 & 0.32 \\
\bottomrule
\end{tabular}
\end{table*}

%% file: tables/supp_tab_frontier.tex
\begin{table}[!htb]
\centering
\caption{\textbf{Accuracy--cost frontier} on 8 ego-and-actor windows at $346\times260$, measured on an exclusive GPU (median of three runs). Event-level F1 (\%) at three tolerances, wall time (s), and rendering cost $Q$. TIDES variants are evaluated without emission gate.}
\label{tab:supp_frontier}
\footnotesize
\setlength{\tabcolsep}{3pt}
\begin{tabular}{lccccc}
\toprule
Configuration & F1@5$\uparrow$ & F1@10$\uparrow$ & F1@31$\uparrow$ & Time$\downarrow$ & $Q\downarrow$\\
\midrule
\method, $\varepsilon_t$=16\,$\mu$s & 99.711 & 99.936 & \fst{99.994} & \fst{2.75} & \fst{0.295} \\
\method, $\varepsilon_t$=8\,$\mu$s & 99.842 & \trd{99.960} & \fst{99.994} & \snd{2.86} & \snd{0.302} \\
\method, $\varepsilon_t$=4\,$\mu$s & \snd{99.861} & \snd{99.963} & \fst{99.994} & \trd{3.31} & \trd{0.314} \\
\method, $\varepsilon_t$=2\,$\mu$s & \fst{99.870} & \fst{99.965} & \fst{99.994} & 4.43 & 0.332 \\
\midrule
Frame 8\,kHz & 99.740 & 99.895 & 99.962 & 3.39 & 4.00 \\
Frame 16\,kHz & 99.800 & 99.910 & 99.985 & 6.81 & 8.00 \\
Frame 32\,kHz & 99.836 & 99.944 & \snd{99.992} & 13.82 & 16.00 \\
Frame 64\,kHz & \trd{99.851} & \trd{99.960} & \fst{99.994} & 28.13 & 32.00 \\
\midrule
Hermite 62.5\,$\mu$s & 99.800 & 99.917 & 99.990 & 10.81 & 8.00 \\
Hermite 31.25\,$\mu$s & 99.826 & 99.946 & \trd{99.991} & 21.49 & 16.00 \\
Hermite 15.625\,$\mu$s & 99.847 & \trd{99.960} & \fst{99.994} & 42.77 & 32.00 \\
\midrule
TIDES-lin. 8\,kHz & 99.352 & 99.791 & 99.928 & 5.96 & 4.03 \\
TIDES-lin. 16\,kHz & 99.771 & 99.912 & 99.968 & 11.01 & 8.03 \\
TIDES-lin. 32\,kHz & 99.816 & 99.932 & 99.980 & 21.31 & 16.02 \\
TIDES-lin. 64\,kHz & 99.829 & 99.938 & 99.983 & 42.11 & 32.02 \\
TIDES-quad. 16\,kHz & 99.822 & 99.938 & 99.988 & 11.41 & 8.03 \\
TIDES-quad. 32\,kHz & 99.834 & 99.947 & \snd{99.992} & 22.34 & 16.02 \\
TIDES-adapt. 8\,kHz & 99.803 & 99.918 & 99.973 & 17.01 & 12.25 \\
TIDES-adapt. 16\,kHz & 99.834 & 99.939 & 99.982 & 33.35 & 24.46 \\
\bottomrule
\end{tabular}
\end{table}

%% file: tables/supp_tab_calibration.tex
\begin{table}[!htb]
\centering
\caption{\textbf{Calibration of the time tolerance} on four ego-and-actor windows (29{,}477 events) against a reference with a 0.98\,$\mu$s step. Median error ($\mu$s), percentage of reference events whose matched event deviates by more than $\varepsilon_t$ and $2\varepsilon_t$ (unmatched events count as deviating), and miss rate (\%) at 5\,$\mu$s.}
\label{tab:supp_calibration}
\footnotesize
\setlength{\tabcolsep}{3pt}
\begin{tabular}{lcccc}
\toprule
Method & Med. $|\Delta t|$ & $>\varepsilon_t$ & $>2\varepsilon_t$ & Miss@5\\
\midrule
\method, $\varepsilon_t$=1\,$\mu$s & 0.29 & 9.48 & 1.74 & 0.15\\
\method, $\varepsilon_t$=2\,$\mu$s & 0.29 & 2.01 & 0.25 & 0.15\\
\method, $\varepsilon_t$=4\,$\mu$s & 0.30 & 0.27 & 0.06 & 0.16\\
\method, $\varepsilon_t$=8\,$\mu$s & 0.31 & 0.07 & 0.02 & 0.22\\
\method, $\varepsilon_t$=16\,$\mu$s & 0.33 & 0.02 & 0.01 & 0.37\\
\midrule
Frame 64\,kHz & 0.24 & -- & -- & 0.17\\
Reference, 3.9\,$\mu$s step & 0.13 & -- & -- & 0.00\\
\bottomrule
\end{tabular}
\end{table}

%% file: tables/supp_tab_m3ed.tex
\begin{table*}[t]
\centering
\caption{\textbf{Generalization to real event recordings on M3ED.} All simulators share the same reconstructed scene of each window. Frame-driven baselines and TIDES use 10$\times$ the camera rate (250\,Hz) following TIDES~\cite{thirgood2026tides}; RGB at the native 25\,Hz is listed for reference. Metrics compare simulated and real events over 24 windows (mean of per-window medians over 3 phase seeds); Avg. rank is the mean rank over these six metrics; the last two columns report fidelity to the scene's own continuous-time limit and the number of full-frame-equivalent renders per 50\,ms window. \colorbox{rkA}{\textbf{First}}, \colorbox{rkB}{second}, and \colorbox{rkC}{third} per column.}
\label{tab:supp_m3ed}
\footnotesize
\setlength{\tabcolsep}{3.3pt}
\begin{tabular}{lc cccccc c c c}
\toprule
& & \multicolumn{6}{c}{Similarity to real events} & & Scene & \\
\cmidrule(lr){3-8}
Method & Rate & Voxel$\uparrow$ & Time surf.$\uparrow$ & Count err.$\downarrow$ & Chamfer$\downarrow$ & log-ISI $W_1\downarrow$ & Fano err.$\downarrow$ & Avg. rank$\downarrow$ & F1@0.25\,ms$\uparrow$ & Renders\\
\midrule
RGB~\cite{gehrig2020vid2e} & 25 Hz & 0.1110 & 0.3160 & 0.466 & 6.30 & 0.507 & 1.549 & 8.33 & 0.104 & 3 \\
RGB~\cite{gehrig2020vid2e} & 250 Hz & \fst{0.1322} & 0.3291 & 0.432 & 6.12 & 0.403 & 1.414 & 4.67 & 0.798 & 16 \\
ESIM~\cite{rebecq2018esim} & 250 Hz & \fst{0.1322} & 0.3291 & 0.432 & 6.12 & 0.403 & 1.414 & 4.67 & 0.798 & 16 \\
v2e~\cite{hu2021v2e} & 250 Hz & 0.1254 & 0.3279 & 0.445 & 6.11 & 0.456 & \snd{1.368} & 5.50 & 0.116 & 16 \\
v2e-HDR~\cite{hu2021v2e} & 250 Hz & 0.1266 & 0.3277 & 0.452 & 6.14 & 0.407 & \trd{1.369} & 6.00 & 0.117 & 16 \\
DVS-Voltmeter~\cite{lin2022dvsvoltmeter} & 250 Hz & 0.0926 & 0.2288 & 0.918 & -- & \fst{0.376} & \fst{0.678} & 5.80 & 0.014 & 16 \\
TIDES-lin.~\cite{thirgood2026tides} & 250 Hz & 0.1297 & \fst{0.3411} & \fst{0.378} & \fst{5.93} & 0.425 & 1.419 & \snd{3.67} & 0.591 & 16 \\
TIDES-quad.~\cite{thirgood2026tides} & 250 Hz & \trd{0.1320} & 0.3311 & 0.423 & 6.11 & \snd{0.390} & 1.389 & \trd{3.83} & \snd{0.904} & 16 \\
TIDES-adapt.~\cite{thirgood2026tides} & adaptive & 0.1313 & \snd{0.3370} & \snd{0.393} & \snd{6.01} & 0.397 & 1.431 & \snd{3.67} & \trd{0.884} & 34 \\
\midrule
\textbf{\method (Ours)} & $\varepsilon_t$=4\,$\mu$s & \snd{0.1321} & \trd{0.3319} & \trd{0.416} & \trd{6.10} & \trd{0.391} & 1.387 & \fst{3.00} & \fst{0.998} & 113 \\
\bottomrule
\end{tabular}
\end{table*}

%% file: tables/supp_tab_m3ed_protocols.tex
\begin{table*}[t]
\centering
\caption{\textbf{M3ED under further protocols}. Native: camera rate; Budget: the same rendered area as \method; Dense: 4\,kHz frames, which approximates the continuous-time limit of the scene. Renders per 50\,ms window. Scene: F1@0.25\,ms to the dense 4\,kHz events of the same scene.}
\label{tab:supp_m3ed_protocols}
\footnotesize
\setlength{\tabcolsep}{4pt}
\begin{tabular}{llcccccc}
\toprule
Protocol & Method & Renders & Voxel$\uparrow$ & T-surf.$\uparrow$ & Count$\downarrow$ & $W_1\downarrow$ & Scene$\uparrow$\\
\midrule
Native (25\,Hz) & Frame & 3 & 0.1110 & 0.3160 & 0.466 & 0.507 & 0.104 \\
 & v2e & 3 & 0.0724 & 0.2717 & 0.692 & 0.558 & 0.022 \\
\midrule
Budget & Frame & 115 & 0.1321 & 0.3319 & 0.417 & 0.391 & 0.999 \\
 & v2e-HDR & 115 & 0.1323 & 0.3320 & 0.416 & 0.392 & 0.472 \\
\midrule
Dense (4\,kHz) & Frame & 241 & 0.1321 & 0.3319 & 0.416 & 0.391 & 1.000 \\
 & v2e-HDR & 241 & 0.1321 & 0.3319 & 0.416 & 0.391 & 1.000 \\
 & DVS-Voltmeter & 241 & 0.0931 & 0.2303 & 0.917 & 0.366 & 0.015 \\
\midrule
\method & $\varepsilon_t$=4\,$\mu$s & 113 & 0.1321 & 0.3319 & 0.416 & 0.391 & 0.998 \\
 & $\varepsilon_t$=62.5\,$\mu$s & 52.5 & 0.1321 & 0.3319 & 0.417 & 0.391 & 0.998 \\
 & $\varepsilon_t$=250\,$\mu$s & 38 & 0.1321 & 0.3319 & 0.417 & 0.391 & 0.998 \\
 & $\varepsilon_t$=1\,ms & 34.5 & 0.1321 & 0.3319 & 0.417 & 0.391 & 0.997 \\
\bottomrule
\end{tabular}
\end{table*}

%% file: tables/supp_tab_fidelity.tex
\begin{table}[!htb]
\centering
\caption{\textbf{Scene fidelity on M3ED} (median over 72 cells). F1 to the 4\,kHz dense events of the same scene at 0.25 and 1\,ms (F1@.25, F1@1), event count relative to the scene limit and to the real recording, and the percentage of events followed by an opposite-polarity event at the same pixel within 2\,ms.}
\label{tab:supp_fidelity}
\footnotesize
\setlength{\tabcolsep}{2.6pt}
\begin{tabular}{lccccc}
\toprule
Method & F1@.25 & F1@1 & Cnt./scene & Cnt./real & Flips (\%)\\
\midrule
\method & 0.998 & 0.999 & 1.000 & 0.678 & 0.00 \\
TIDES-quad. & 0.904 & 0.986 & 0.999 & 0.677 & 0.00 \\
TIDES-adapt. & 0.884 & 0.953 & 1.013 & 0.757 & 0.07 \\
Frame 250\,Hz & 0.798 & 0.952 & 0.993 & 0.674 & 0.00 \\
TIDES-lin. & 0.591 & 0.832 & 1.056 & 0.811 & 0.57 \\
v2e 250\,Hz & 0.116 & 0.361 & 0.966 & 0.650 & 0.00 \\
Real events & -- & -- & -- & 1.000 & 1.98 \\
\bottomrule
\end{tabular}
\end{table}

%% file: tables/supp_tab_flow.tex
\begin{table}[!htb]
\centering
\caption{\textbf{Sim-to-real optical flow}, all training inputs (mean over 5 seeds). Test: end-point error (px) and percentage of pixels with error above 3\,px; dev-test EPE is diagnostic only. Budget: frame-driven simulation with the same rendered area as \method.}
\label{tab:supp_flow}
\footnotesize
\setlength{\tabcolsep}{2.8pt}
\begin{tabular}{llccc}
\toprule
Protocol & Training & Test EPE$\downarrow$ & $>$3\,px$\downarrow$ & Dev EPE$\downarrow$\\
\midrule
Controls & No events & $2.533\pm0.055$ & 16.5 & 1.155 \\
 & Shuffled events & $2.576\pm0.035$ & 17.6 & 1.355 \\
\midrule
25\,Hz & Frame & $1.543\pm0.078$ & 12.4 & 0.810 \\
 & ESIM & $1.556\pm0.128$ & 12.6 & 0.816 \\
 & v2e & $2.154\pm0.071$ & 13.3 & 0.844 \\
 & v2e-HDR & $2.110\pm0.158$ & 13.4 & 0.842 \\
\midrule
250\,Hz & Frame & $0.993\pm0.058$ & 5.1 & 0.844 \\
 & ESIM & $1.048\pm0.064$ & 6.5 & 0.839 \\
 & v2e & $0.984\pm0.036$ & 4.8 & 0.831 \\
 & v2e-HDR & $1.027\pm0.051$ & 5.2 & 0.833 \\
 & DVS-Voltmeter & $1.549\pm0.122$ & 9.2 & 1.517 \\
 & TIDES-lin. & $1.008\pm0.030$ & 5.6 & 0.792 \\
 & TIDES-quad. & $1.009\pm0.048$ & 5.2 & 0.763 \\
 & TIDES-adapt. & $1.012\pm0.032$ & 5.3 & 0.833 \\
\midrule
Budget & Frame & $1.026\pm0.090$ & 5.9 & 0.798 \\
 & ESIM & $1.073\pm0.026$ & 6.3 & 0.812 \\
 & v2e & $1.016\pm0.093$ & 4.8 & 0.833 \\
 & v2e-HDR & $0.973\pm0.057$ & 4.6 & 0.813 \\
\midrule
4\,kHz & Frame & $1.043\pm0.043$ & 6.1 & 0.819 \\
 & ESIM & $1.028\pm0.071$ & 5.5 & 0.862 \\
 & v2e & $1.013\pm0.055$ & 5.6 & 0.850 \\
 & v2e-HDR & $1.045\pm0.029$ & 5.9 & 0.824 \\
 & DVS-Voltmeter & $1.573\pm0.162$ & 10.3 & 1.594 \\
\midrule
Ours & \method & $1.032\pm0.060$ & 6.3 & 0.799 \\
\midrule
Real & Real events & $1.008\pm0.043$ & 5.0 & 0.744 \\
\bottomrule
\end{tabular}
\end{table}

%% file: tables/supp_tab_long.tex
\begin{table}[!htb]
\centering
\caption{\textbf{Long sequences} (5\,s, $346\times260$). Gaussians and events in millions, mean and 95th-percentile cost $Q$ over base intervals, peak GPU memory in GB, wall time in minutes, and unresolved intervals. Wall times were recorded with several runs sharing each GPU and are therefore upper bounds. nuScenes 189 is nearly static.}
\label{tab:supp_long}
\footnotesize
\setlength{\tabcolsep}{3pt}
\begin{tabular}{lccccccc}
\toprule
Scene & Gauss. & Events & $Q$ & $Q_{95}$ & Mem. & Time & Unres.\\
\midrule
PandaSet 011 & 1.61 & 5.11 & 0.46 & 1.22 & 5.5 & 102 & 0 \\
PandaSet 024 & 1.84 & 3.56 & 0.38 & 1.03 & 9.9 & 29 & 0 \\
PandaSet 032 & 1.87 & 7.05 & 0.53 & 1.16 & 10.1 & 52 & 0 \\
PandaSet 058 & 1.05 & 2.54 & 0.52 & 1.27 & 8.1 & 80 & 0 \\
PandaSet 068 & 0.92 & 5.84 & 0.50 & 1.13 & 7.8 & 44 & 0 \\
PandaSet 086 & 2.15 & 1.49 & 0.37 & 1.14 & 6.7 & 80 & 0 \\
PandaSet 095 & 2.06 & 5.65 & 0.48 & 1.08 & 7.0 & 75 & 0 \\
PandaSet 120 & 2.00 & 3.34 & 0.42 & 1.14 & 6.6 & 119 & 0 \\
\midrule
nuScenes 025 & 1.00 & 2.89 & 0.38 & 1.17 & 3.7 & 28 & 0 \\
nuScenes 034 & 1.19 & 8.08 & 0.48 & 1.05 & 4.4 & 25 & 0 \\
nuScenes 050 & 1.84 & 1.97 & 0.51 & 1.21 & 8.3 & 92 & 0 \\
nuScenes 088 & 1.17 & 6.52 & 0.53 & 1.17 & 6.4 & 103 & 0 \\
nuScenes 176 & 1.43 & 4.79 & 0.56 & 1.24 & 5.2 & 133 & 0 \\
nuScenes 189 & 1.45 & 0.00 & 0.26 & 1.00 & 2.7 & 2 & 0 \\
nuScenes 300 & 0.80 & 1.20 & 0.41 & 1.22 & 6.4 & 25 & 0 \\
nuScenes 407 & 0.92 & 3.15 & 0.40 & 1.11 & 5.3 & 47 & 0 \\
\bottomrule
\end{tabular}
\end{table}